\documentclass[journal]{IEEEtran}
\usepackage{threeparttable}
\usepackage{graphicx}
\usepackage{amsmath}
\usepackage{amsfonts}
\usepackage{multirow}
\usepackage{multicol}
\usepackage{cite}
\usepackage[numbers]{natbib}
\usepackage{setspace}
\usepackage[ruled,linesnumbered]{algorithm2e}
\usepackage{color}
\usepackage{subfig}
\usepackage{bbding}
\ifCLASSINFOpdf
  
\else
 
\fi

\begin{document}

\title{Stacked BNAS: Rethinking Broad Convolutional Neural Network for Neural Architecture Search
\thanks{This work has been submitted to the IEEE for possible publication. Copyright may be transferred without notice, after which this version may no longer be accessible.}}

\author{Zixiang~Ding,
        Yaran~Chen,~\IEEEmembership{Member,~IEEE},
        Nannan~Li,
        Dongbin~Zhao,~\IEEEmembership{Fellow,~IEEE},
        and~C.L.Philip~Chen,~\IEEEmembership{Fellow,~IEEE}
\thanks{Z. Ding, Y. Chen, N. Li and D. Zhao are with  the State Key Laboratory of Management and Control for Complex Systems, Institute of Automation, Chinese Academy of Sciences, Beijing 100190, China and also with the School of Artificial Intelligence, University of Chinese Academy of Sciences, Beijing 100049, China (email : dingzixiang2018@ia.ac.cn, chenyaran2013@ia.ac.cn, linannan2017@ia.ac.cn, dongbin.zhao@ia.ac.cn).}
\thanks{C. L. P. Chen is with the School of Computer Science \& Engineering, South China University of Technology, Guangzhou, Guangdong 510006, China, and also with the College of Navigation, Dalian Maritime University, Dalian 116026, China (e-mail: philip.chen@ieee.org).}
}


\markboth{IEEE Transactions on Systems, Man, and Cybernetics: Systems, ~2022}%
{Shell \MakeLowercase{\textit{et al.}}: Bare Demo}

\maketitle

\begin{abstract}
Different from other deep scalable architecture-based NAS approaches, Broad Neural Architecture Search (BNAS) proposes a broad scalable architecture which consists of convolution and enhancement blocks, dubbed Broad Convolutional Neural Network (BCNN), as the search space for amazing efficiency improvement. BCNN reuses the topologies of cells in the convolution block so that BNAS can employ few cells for efficient search. Moreover, multi-scale feature fusion and knowledge embedding are proposed to improve the performance of BCNN with shallow topology. However, BNAS suffers some drawbacks: 1) insufficient representation diversity for feature fusion and enhancement and 2) time consumption of knowledge embedding design by human experts. This paper proposes Stacked BNAS,  whose search space is a developed broad scalable architecture named Stacked BCNN, with better performance than BNAS. On the one hand, Stacked BCNN treats mini BCNN as a basic block to preserve comprehensive representation and deliver powerful feature extraction ability.  For multi-scale feature enhancement, each mini BCNN feeds the outputs of deep and broad cells to the enhancement cell. For multi-scale feature  fusion, each mini BCNN feeds the outputs of deep, broad and enhancement cells to the output node. On the other hand, Knowledge Embedding Search (KES) is proposed to learn appropriate knowledge embeddings in a differentiable way. Moreover, the basic unit of KES is an over-parameterized knowledge embedding module that consists of all possible candidate knowledge embeddings.  Experimental results show that 1) Stacked BNAS obtains better performance than BNAS-v2 on both CIFAR-10 and ImageNet, 2) the proposed KES algorithm contributes to reducing the parameters of the learned architecture with satisfactory performance, and 3) Stacked BNAS delivers a state-of-the-art efficiency of  0.02 GPU days. 
\end{abstract}


\IEEEpeerreviewmaketitle

\section{Introduction}

\IEEEPARstart{A}{rtificial} neural networks have shown powerful capability of feature extraction in various fields, e.g., classification \cite{wen2020multilabel}, detection \cite{tao2020detection}. To design high-performance neural networks automatically, Neural Architecture Search (NAS) is proposed. NAS has achieved unprecedented accomplishments for various structures (e.g., convolutional neural network \cite{zoph2017neural}, tensor ring \cite{li2021heuristic}, language model \cite{pham2018efficient}) design. However, it needs enormous computational requirements, e.g., more than 22000 GPU days for vanilla NAS \cite{zoph2017neural}. NASNet \cite{zoph2018learning} proposes cell search space to alleviate the time-consuming issue and delivers a higher efficiency of 1800 GPU days. However, the search cost is unbearable yet. Subsequently, weight-sharing NAS pipeline is proposed to deliver novel efficiency. Reinforcement Learning (RL) based ENAS \cite{pham2018efficient} needs only 0.45 GPU days via parameter sharing. Gradient-based DARTS \cite{liu2018darts} employs a continuous relaxation strategy to transfer the search space from discrete to continuous, and delivers a efficiency of 0.45 GPU days. Furthermore, based on the DARTS framework, PC-DARTS \cite{xu2019pc} delivers a state-of-the-art search efficiency of 0.1 GPU days via partial channel connections.  

\begin{figure}[t]
\centering
\includegraphics[width=0.45\textwidth]{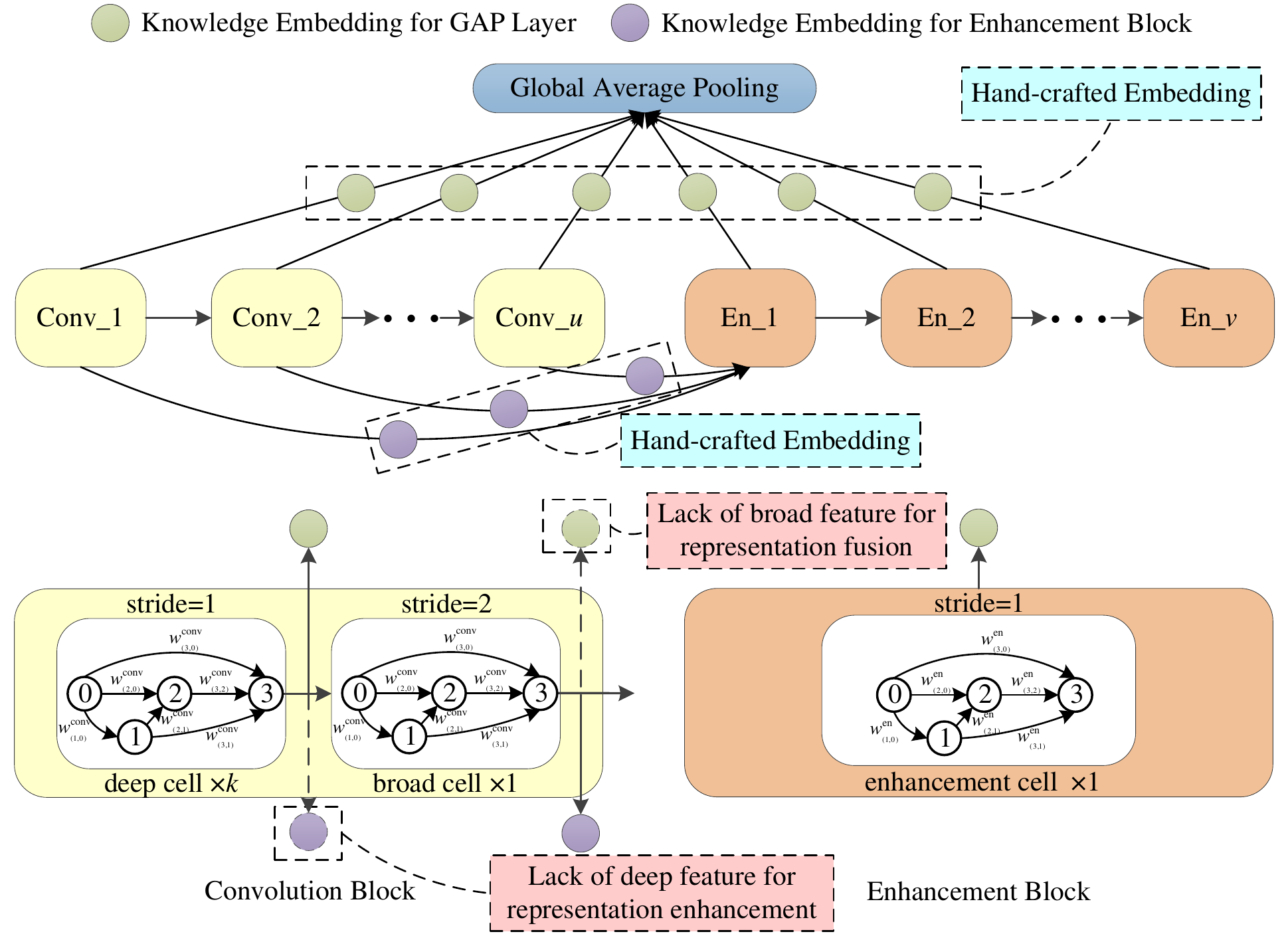}
\captionsetup{font={small}} 
\caption{Issues of BCNN. 1) Lack of representation diversity for feature fusion and enhancement: only deep and broad feature information is fed into the GAP layer for representation fusion and  the first enhancement block for representation enhancement, respectively. 2) It is time-consuming to design knowledge embeddings by experts.}
\label{fig::issues}
 \vspace{-0.5cm}
\end{figure}

Different from deep search space based approaches, BNAS \cite{ding2021bnas} proposed a broad search space named Broad Convolutional Neural Network (BCNN). Benefiting of BCNN, RL-based BNAS delivers an efficiency of 0.2 GPU days, which is 2.2$\times$ faster than ENAS \citep{pham2018efficient}. However, the training mechanism of BNAS following ENAS leads to terrible unfair training issue \cite{chu2019fairnas}. BNAS-v2 \cite{ding2021bnas-v2} was proposed to solve the above issue by a differentiable broad search space with a efficiency of 0.05 GPU days. Admittedly, BNASs achieve satisfactory performance, especially in terms of efficiency. Nevertheless, BCNN suffers two drawbacks as shown in Fig.  \ref{fig::issues}: 1) insufficient representation diversity for feature fusion and enhancement and 2) time-consuming knowledge embedding design.  
  
This paper proposes Stacked BNAS whose scalable architecture is named Stacked BCNN, which treats mini BCNN as its basic block. Moreover, each mini BCNN can feed sufficient representations to the GAP layer and enhancement block for feature fusion and enhancement, respectively. As a new paradigm of neural networks, this paper prove also the universal approximation ability of Stacked BCNN. Furthermore, the knowledge embedding design task is transferred from discrete to continuous space, and learn appropriate knowledge embeddings in a differentiable way to solve the second issue. Our contributions can be summarized as follows:
\begin{itemize}
\item \textbf{Stacked BNAS:} Stacked BNAS is proposed to further improve the performance of NAS via Stacked BCNN. 
\item \textbf{Stacked BCNN:} This paper does not only propose a new broad search space dubbed Stacked BCNN for NAS, but also mathematically analyze the universal approximation ability of the proposed Stacked BCNN. 
\item \textbf{Knowledge Embedding Search:} An algorithm is also proposed for knowledge embedding design.
\item \textbf{Efficiency:} Contributing to the proposed Stacked BCNN and optimization algorithm, Stacked BNAS delivers a state-of-the-art efficiency of 0.02 days on a single NVIDIA GTX 1080Ti GPU.   
\end{itemize}

\section{Related Work}
\label{sec::related_work}

\subsection{Neural Architecture Search}

\citet{elsken2018neural} claimed that NAS consists of three components: search space, optimization strategy and estimation strategy. The search space referred to not only primitive operators, but also the combination paradigm of those candidate operations \cite{he2021automl}. As described in \cite{zoph2017neural}, there were mainly five types of search spaces: entire-structured, cell-based \cite{zhong2018practical}, hierarchical \cite{liu2019auto}, morphism-based \cite{wei2016network} and broad \cite{ding2021bnas-v2}. These search spaces are briefly introduced as below.


\subsubsection{Entire-structured Search Space}

In the entire-structured search space, each layer with a specified operation (e.g., various convolutions and average pooling) was stacked one after another \cite{zoph2017neural}. Beyond that, the skip connection was used in the above search space to explore more complex neural architectures. This search space had two disadvantages: computationally expensive and insufficient transferability \cite{he2021automl}.  

\subsubsection{Cell-based Search Space}

To mitigate the above issues of the entire-structured search space, \citeauthor{zoph2018learning} (\citeyear{zoph2018learning}) proposed a cell-based search space where each cell with a list of operations was stacked to construct a deep search space. The cell-based search space consists of normal and reduction cells that have different architectures and strides. Moreover, each cell treats the outputs of two predecessors as its inputs. Due to the effectiveness of cell-based search space in terms of efficiency and transferability, many cell-based NAS approaches were proposed, e.g., weight-sharing ENAS \cite{pham2018efficient}, differentiable DARTS \cite{liu2018darts}. 

\subsubsection{Hierarchical Search Space}

Most cell-based NAS approaches \cite{pham2018efficient,liu2018darts} followed a two-level hierarchy: the inner cell level and the outer network level. On the one hand, the inner level chose operation and connection of each intermediate node in the cell. On the other hand, the outer level controlled the spatial-resolution changes. A general formulation \cite{liu2019auto} was proposed to learn the network-level structure. \citet{liu2018hierarchical} proposed a hierarchical architecture representation to avoid manually predefining the network block number.    

\subsubsection{Morphism-based Search Space}

The morphism-based search space directly modified the existing architecture. Net2Net \cite{chen2015net2net} employed identity morphism (IdMorph) to design architecture based on the existing model.  \citet{he2016deep} claimed that there were several issues in IdMorph: 1) limited width and depth changes and 2) identity layer drawback. To solve the above issue, a developed approach named network morphism \cite{wei2016network} was proposed. Network morphism adopted a  parameter sharing strategy \cite{pham2018efficient} to inherit the knowledge from the parent network to the child network, which grew into a more robust model. Furthermore, network morphism was improved to deliver better performance in terms of optimization algorithm \cite{jin2019auto} and its level \cite{wei2017modularized}. Furthermore, \citet{chen2020modulenet} used hand-crafted and learned blocks to discover novel architecture via parameter  inheriting.    

\subsubsection{Broad Search Space}

\begin{figure}[!h]
  \centering
    \subfloat[The structure of BCNN]{\includegraphics[width=0.46\textwidth]{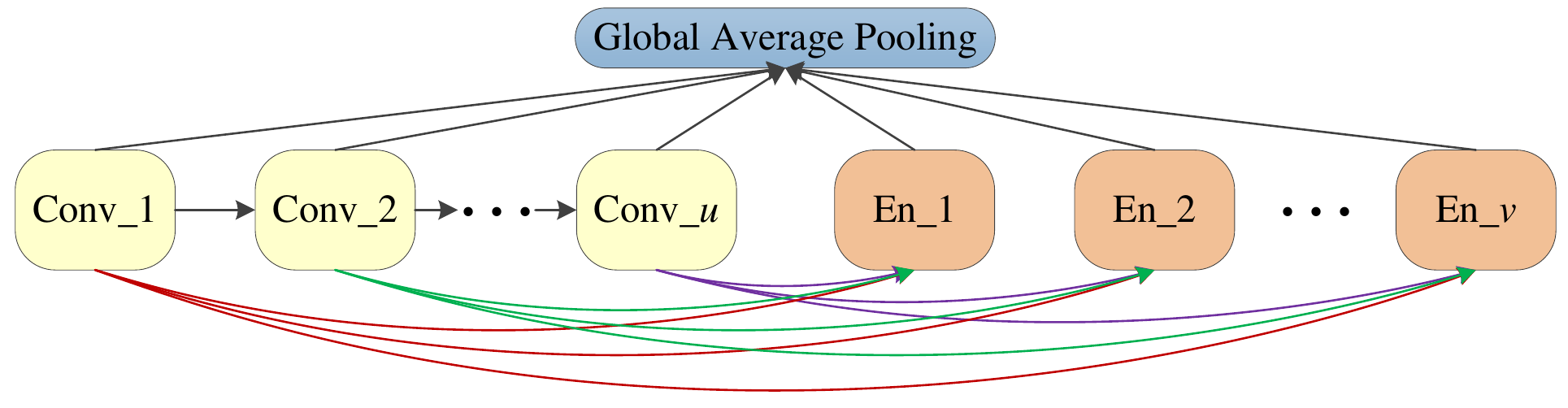}} \\
    \subfloat[The structure of BCNN-CCLE]{\includegraphics[width=0.46\textwidth]{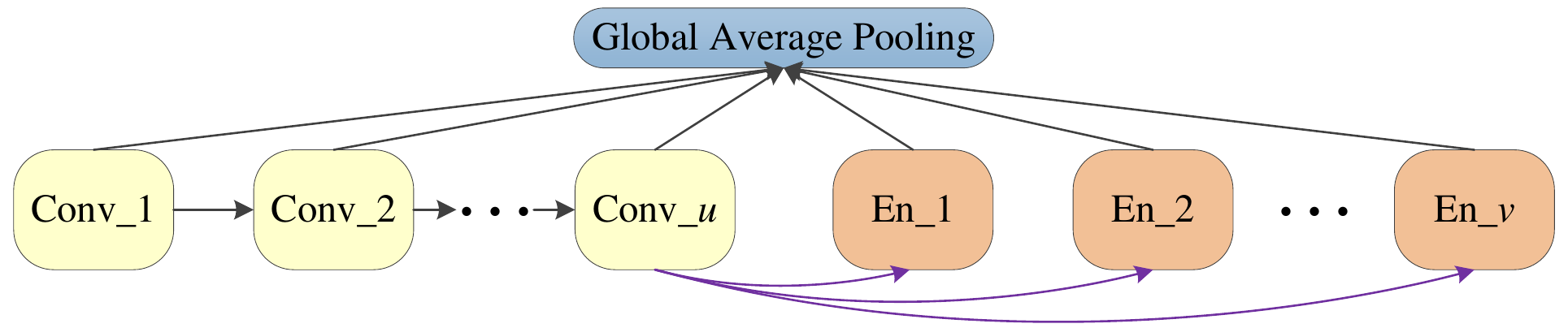}} \\
\captionsetup{font={small}}   
  \caption{Broad search space \cite{ding2021bnas}.}
    \label{fig::bcnns}
     \vspace{-0.5cm}
\end{figure}

\citet{ding2021bnas} proposed a broad search space named BCNN, which employs broad topology to obtain extreme fast search speed and satisfactory performance. BCNN belongs to the cell-based network-level search space. There are three broad search spaces: BCNN (see Fig. \ref{fig::bcnns} (a)), BCNN-CCLE (see Fig. \ref{fig::bcnns} (b)) and BCNN-CCE (see Fig. \ref{fig::issues}). BCNNs consist of convolution and enhancement blocks which densely connect with the GAP layer for multi-scale feature fusion. The main difference among BCNNs is the connection between convolution and enhancement blocks. Similarly, \citeauthor{fang2020densely} (\citeyear{fang2020densely}) proposed a network-level deep search space named dense search space where the MBConv \cite{sandler2018mobilenetv2} is treated as its basic block rather than cell. In contrast, each block of dense search space connects to several subsequent blocks. Based on the broad search space, the efficiency of BNAS \cite{ding2021bnas} was 0.2 GPU days via RL. However, BNAS suffers from unfair training issue, so its optimization mechanism does not take full advantage of the BCNN, i.e., the efficiency improvement is limited. Furthermore, a differentiable over-parameterized broad search space was proposed to solve the unfair training issue in BNAS-v2 \cite{ding2021bnas-v2}. Experimental results show that BNAS-v2 can deliver 2$\times$ faster search efficiency than BNAS when BCNN's all advantages works.  

\subsection{Broad Learning System (BLS)}

Inspired by Random Vector Functional
Link Neural Network (RVFLNN) \citep{pao1992functional} and incremental learning strategy \citep{chen1999rapid}, \citet{chen2017broad} proposed BLS and several variants \citep{chen2018universal}. Subsequently, BLSs were widely used in many fields, e.g., image classification \citep{ zhao2020semi}, industrial control \citep{chu2019weighted}. 

\begin{figure}[h]
\centering
\includegraphics[width=0.43\textwidth]{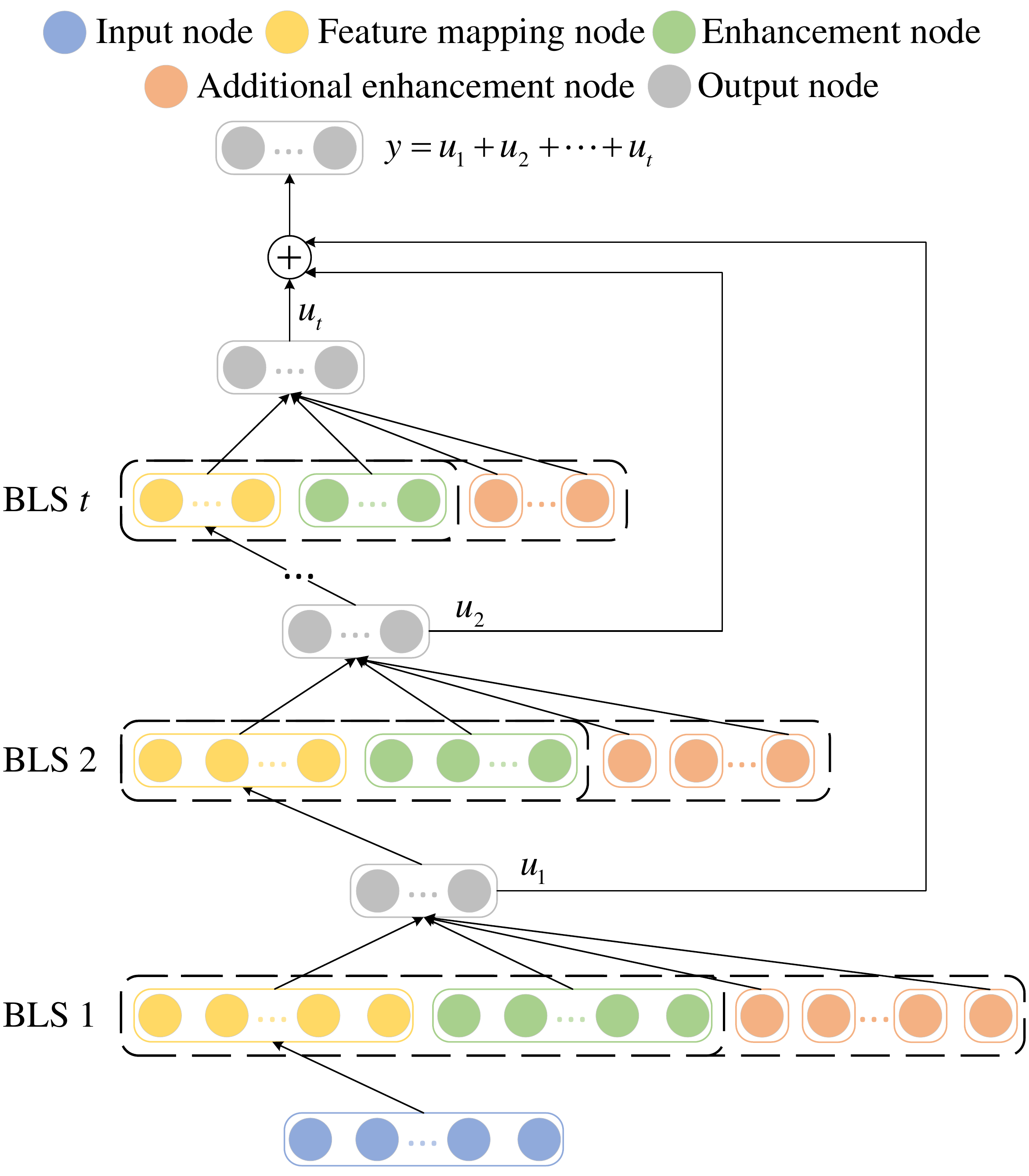}
\captionsetup{font={small}} 
\caption{The structure of Stacked BLS \citep{liu2020stacked} with $i$ BLSs.}
\label{fig::stacked_bls}
 \vspace{-0.5cm}
\end{figure}

To combine the superiority of deep model and BLS, \citet{liu2020stacked} proposed Stacked BLS, whose structure is shown in Fig. \ref{fig::stacked_bls}. BLS was the basic block of Stacked BLS whose output was the combination of all BLSs' outputs. \citet{liu2020stacked} claimed that Stacked BLS only efficiently optimized trainable weights via an incremental learning algorithm, but also extracted deep representation using multiple BLSs.      

\section{Stacked Broad Neural Architecture Search}
\label{sec::methodology}


In this section, a developed broad search space named Stacked BCNN is first proposed to solve the first issue of vanilla BCNN, i.e., insufficient presentation diversity for feature fusion and enhancement. Then, Knowledge Embedding Search (KES) algorithm is proposed to solve the second issue of vanilla BCNN, i.e the time consumption of knowledge embedding design by human experts. Finally, the optimization algorithm of Stacked BNAS is given.  

\subsection{Search Space: Stacked BCNN}
\label{sec::ss}

To improve the feature diversity, mini BCNN based Stacked BCNN is proposed. The structures of Stacked BCNN and mini BCNN are shown in Fig. \ref{fig::Stacked_BCNN}. 

\subsubsection{Stacked BCNN}

\begin{figure}[t]
\centering
\captionsetup{font={small}} 
\subfloat[Stacked BCNN]{\includegraphics[width=0.41\textwidth]{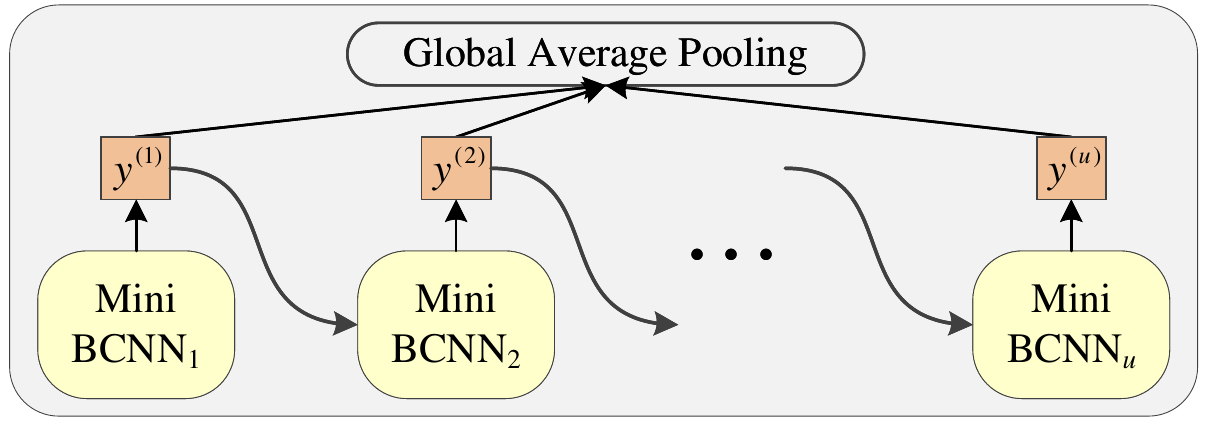}} \\
\subfloat[Mini BCNN]{\includegraphics[width=0.49\textwidth]{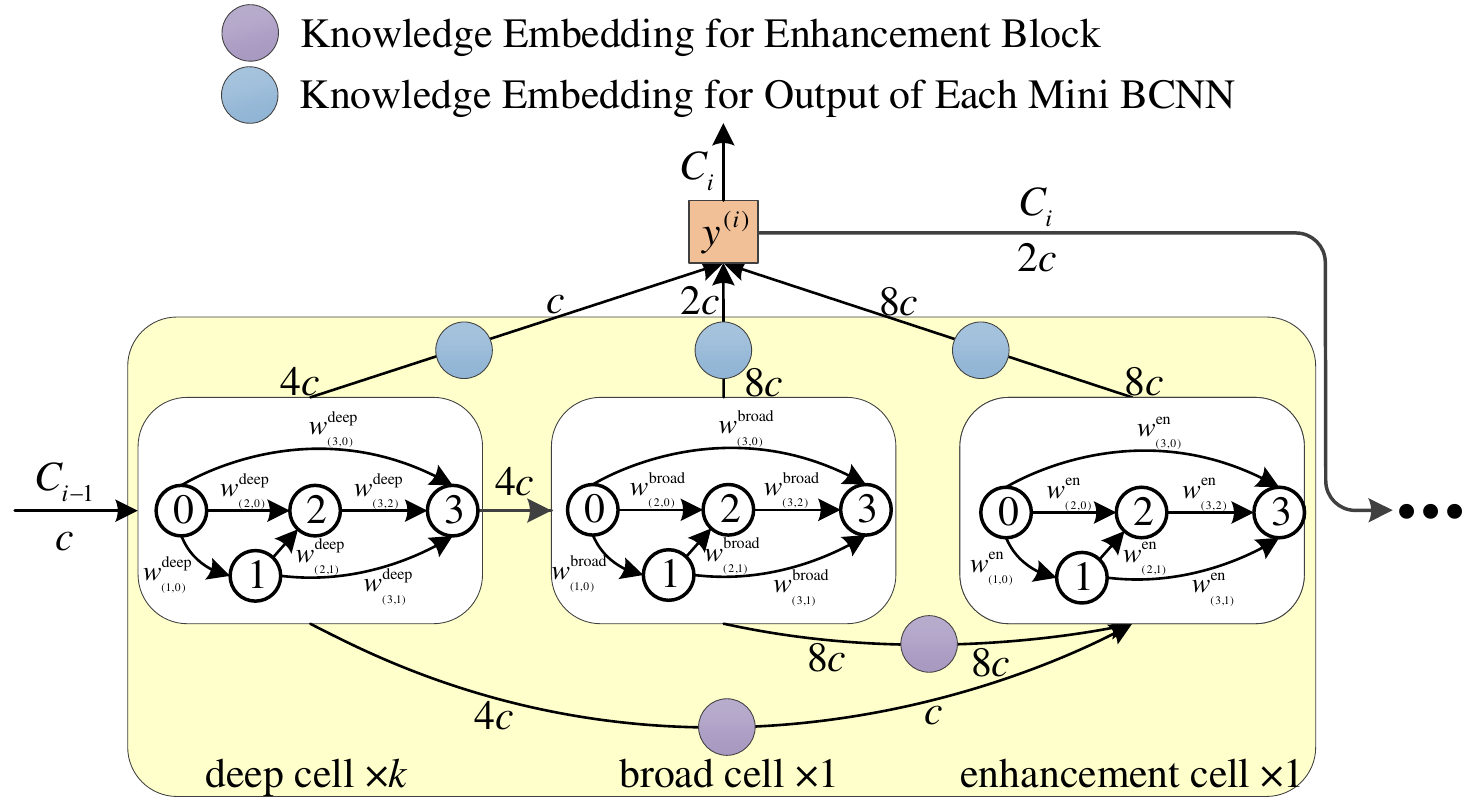}} \\
\caption{Search space of Stacked BNAS. The output of $(i-1)$-th mini BCNN $C_{i-1}$ whose channels are set to $c$, is treated as the inputs for $i$-th mini BCNN. The output channels of each deep, broad and enhancement cells are $4c$, $8c$ and $8c$, respectively. Subsequently, they are fed into the knowledge embeddings whose output channels are $c$, $2c$ and $8c$ for the outputs of deep, broad and enhancement cells, respectively. For the GAP layer and $(i+1)$-th mini BCNN, the output channels of $i$-th mini BCNN are $2c$ except the last one which feeds its all channels to the GAP layer.}
\label{fig::Stacked_BCNN}
 \vspace{-0.5cm}
\end{figure}

As shown in Fig. \ref{fig::Stacked_BCNN} (a), the proposed Stacked BCNN consists of $u$ mini BCNNs, where $u$ is determined by the input size of the first mini BCNN (mini BCNN$_1$). To preserve the multi-scale feature fusion ability of the vanilla BCNN, the Stacked BCNN feed the output of each mini BCNN into GAP layer. 

\subsubsection{Mini BCNN}

Fig. \ref{fig::Stacked_BCNN} (b) shows the structure of mini BCNN. Mini BCNN consists of $k+1$ convolution cells ($k$ 1-stride deep cells, single 2-stride broad cell) and 1 enhancement cell with stride 1. Deep and broad cells are used for deep and broad feature extraction, respectively. The enhancement cell treats both the outputs of deep and broad cells as inputs to obtain a comprehensive enhancement representation. Furthermore, a family of $1\times 1$ convolutions is inserted into fixed locations as the knowledge embeddings. 

There are two main differences between BCNN and Stacked BCNN: 1) the output of each mini BCNN is the combination of all outputs of three cells rather than the outputs of deep and enhancement cells, and 2) the enhancement cell treats the outputs of convolution cells as inputs. The above two differences provide sufficient feature diversity to the GAP layer and enhancement block for representation fusion and enhancement so that better performance can be obtained.  

\subsubsection{Mathematical Information Flow} 
The Stacked BCNN employs a $3\times3$ convolution as the stem layer, and its output is treated as the two inputs of the first deep cell of mini BCNN$_1$, denoted as $y^{(1)}_{-1}$ and $y^{(1)}_{0}$. Similarly, we treat the output of mini BCNN$_i$ as the two inputs of mini BCNN$_{i+1}$.

For mini BCNN$_i (i=1,2,\dots,u)$, its output $y^{(i)}$ can be obtained by the outputs of three cells: 
\begin{align}
y^{(i)}=\phi(\delta_{do}^{(i)}(y^{(i)}_{k}), \delta_{bo}^{(i)}(y^{(i)}_{k+1}), \delta_{eo}^{(i)}(y^{(i)}_{k+2})), \label{eq::mini_bcnn}
\end{align}
where, $\phi$ and $\delta_{*o}^{(i)}$ are concatenation of the channel dimension and knowledge embeddings with respect to mini BCNN$_i$'s output, respectively. Moreover, $y^{(i)}_{k}$ is the output of the last deep cell with a list of operations $\varphi_d$ and can be computed by 
\begin{align}
y^{(i)}_{k}=\varphi_{d}(y^{(i)}_{k-2}, y^{(i)}_{k-1}; {\pmb{W}^{(i)}_d, \pmb{\theta}^{(i)}_d)}, \label{deep}
\end{align}  
$y^{(i)}_{k+1}$ is the output of a broad cell with a list of operations $\varphi_b$ that can be obtained by
\begin{align}
y^{(i)}_{k+1}=\varphi_{b}(y^{(i)}_{k-1}, y^{(i)}_{k}; {\pmb{W}^{(i)}_b, \pmb{\theta}^{(i)}_b)}, \label{broad}
\end{align}
and $y^{(i)}_{k+2}$ is the output of the enhancement cell with a list of operations $\varphi_e$ that can be calculated by
\begin{align}
y^{(i)}_{k+2}=\varphi_{e}(\delta_{de}^{(i)}(y^{(i)}_{k}), \delta_{be}^{(i)}(y^{(i)}_{k+1}); {\pmb{W}^{(i)}_e, \pmb{\theta}^{(i)}_e)}, \label{en}
\end{align}
where, $\delta_{*e}^{(i)}$ represents knowledge embeddings with respect to the enhancement cell of mini BCNN$_i$ and $\pmb{W}^{(i)}_{*}$ and $\pmb{\theta}^{(i)}_{*}$ are the weight and bias matrices of corresponding cells in mini BCNN$_i$, respectively.  

The output of Stacked BCNN can be computed by
\begin{align}
\pmb{y}=\phi(y^{(1)}, y^{(2)}, \dots, y^{(u)}).
\label{eq::output}
\end{align}

\subsubsection{Channel Flow Graph}

As shown in Fig. \ref{fig::Stacked_BCNN} (a), for mini BCNN$_i$, the channel number of the deep cell's output $c_{deep}^{(i)}$ can be obtained by 
\begin{align}
c_{deep}^{(i)}=N_{in} \times 2^{i-1} \times c, \quad i=1,2,\dots,u 
\end{align}
where $N_{in}$ represents the pre-defined intermediate node number of the cell, and $c$ is the input channel number of mini BCNN$_1$. The channel numbers of broad and enhancement cells' outputs in mini BCNN$_i$ are both $2 \times c_{deep}^{(i)}$. For those direct-connected cells and output nodes, corresponding knowledge embedding does not reduce the feature significance. In contrast, the significance of the indirect-connected features is reduced by a factor of a quarter, as shown in Fig. \ref{fig::embedding_search} (a).

\subsection{Knowledge Embedding Search}
\label{sec::embedding}

Different from vanilla BNAS, Knowledge Embedding Search (KES) algorithm which treats an over-parameterized knowledge embedding module as a basic unit, is proposed to learn appropriate knowledge embedding in a differentiable way rather than designing by human. 

\subsubsection{Over-parameterized Knowledge Embedding Module}

\begin{figure}[h]
\centering
\includegraphics[width=0.45\textwidth]{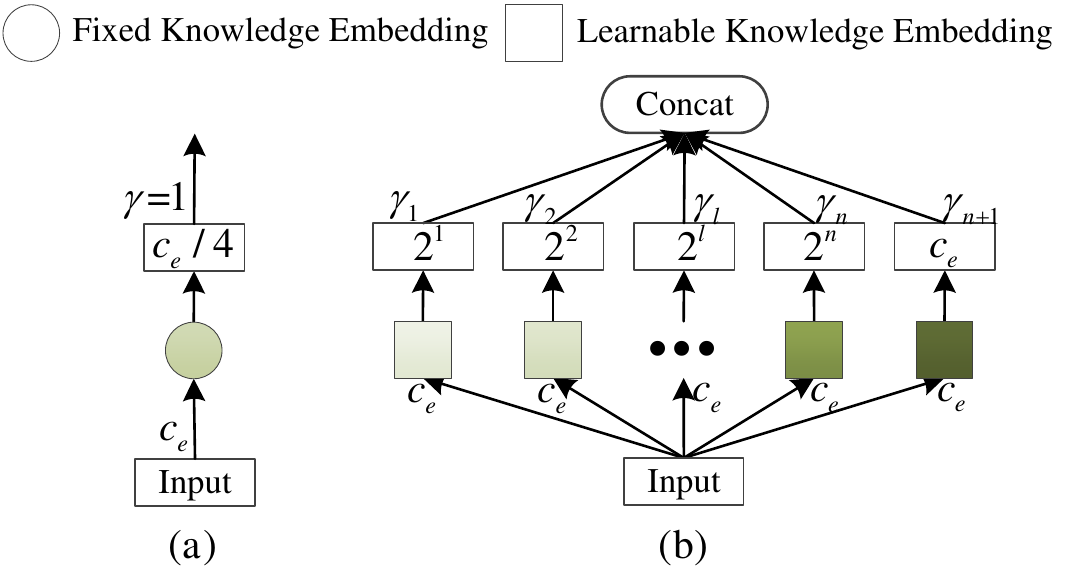}
\captionsetup{font={small}} 
\caption{Embedding between indirectly connected cells and the output node. (a) Hand-crafted knowledge embedding and (b) over-parameterized knowledge embedding module.}
\label{fig::embedding_search}
 \vspace{-0.5cm}
\end{figure}

For each knowledge embedding on an indirect edge, an over-parameterized knowledge embedding module, as shown in  Fig. \ref{fig::embedding_search} (b), is constructed to discover appropriate knowledge embeddings. Moreover, for the case of $2^n$==$c_e$, we employ two $1\times1$ convolutions with $c_e$ output channels. 

There is an assumption that $c_e$ channels are fed into the indirect-connected knowledge embedding. The over-parameterized knowledge embedding consists of $n$ learnable knowledge embeddings with $2^i (i=1,2,\dots,n)$ output channels and a single learnable knowledge embedding with $c_{e}$ output channels, where $n$ is the largest power of 2 less than $c_{e}$ determined by
\begin{align}
\mathop{{\rm argmax}}\limits_{n} \; 2^n \quad {\rm s.t.} \; 2^n<c_{e}.
\label{eq::embedding} 
\end{align}
Subsequently, the output of the over-parameterized knowledge embedding module $y_{e}$ is obtained by the channel-dimension concatenation of weighted $n+1$ learnable knowledge embedding outputs as
\begin{align}
y_{e} = \phi(\gamma_{1}y^{(1)}_e, \gamma_{2}y^{(2)}_e, \dots, \gamma_{n+1}y^{(n+1)}_e), 
\label{eq::embedding_concat}
\end{align}
where, $y^{(l)}_e$ and $\gamma_l$  $(l=1,2,\dots,n+1)$ represent the weighted output and weight of $l$-th learnable knowledge embedding, respectively. 

\subsubsection{Learning Strategy}

After over-parameterized knowledge embedding module construction, Stacked BNAS aims to jointly optimize the knowledge embedding weights $\gamma$ and network weights $w$. The goal of KES is to discover $\gamma^*$ that minimizes the validation loss $\mathcal{L}_{val}(w^*,\gamma^*)$, where $w^*$ is obtained by minimizing the training loss $\mathcal{L}_{train}(w, \gamma^*)$. The bilevel optimization problem with lower-level variable $w$ and upper-level variable $\gamma$ can be represented as
\begin{align}
\begin{aligned}
\mathop{\rm{min}}\limits_{\gamma}\quad & \mathcal{L}_{val}(w^*(\gamma), \gamma), \\
\rm{s.t.}\quad & w^*(\gamma) = \mathop{\rm{argmin}}\limits_w \ \mathcal{L}_{train}(w, \gamma). 
\end{aligned}
\label{eq::embedding_search}
\end{align}   


\begin{algorithm}[!t]
\caption{Stacked BNAS}
\small
\label{algorithm}
Define $p$ as the early stopping epoch number, $q$ as the stable epoch number with zero initialization, $kes$ as the flags represented using KES or not, $arch_{prev}$ as previous architecture with None initialization\;
For each edge $(i,j)$, use \eqref{eq::pc} and \eqref{xpc} for continuous relaxation as $f^{PC}_{(i,j)}(x_{(j)}^{PC}; M_{(i,j)})$ parameterized by $\alpha_{(i,j)}$ and $\beta_{(i,j)}$\;
Define architecture weight set $\Theta = [\alpha, \beta]$\;
\If {$kes$}{ Use \eqref{eq::embedding} and \eqref{eq::embedding_concat} to construct over-parameterized knowledge embedding module parameterized by $\gamma$\;
$\Theta = [\alpha, \beta, \gamma]$\;}
\While{not converged}
{Optimize $w$ by descending $\nabla_{w}\mathcal{L}_{train}(w,\Theta)$\;
Optimize $\Theta$ by descending $\nabla_{\Theta}\mathcal{L}_{val}(w-\xi\nabla_{w}\mathcal{L}_{train}(w,\Theta),\Theta)$\;
Determine current architecture $arch_{curr}$ by taking {\rm argmax}\;
$arch_{prev}=arch_{curr}$\;
\eIf{$arch_{curr} == arch_{prev}$}
{$q = q+1$\;
\If {$q \geq p$}{break}}
{$q = 1$\;}}
Output $arch_{curr}$ as the best architecture.
\end{algorithm}

The optimization of \eqref{eq::embedding_search} exactly is prohibitive because $w^*(\gamma)$ needs to be recomputed by the second term of \eqref{eq::embedding_search} whenever $\gamma$ takes place any change \cite{liu2018darts}. Therefore, an approximate iterative optimization process is proposed as follows. For each gradient descent step,  network weights $w$ and knowledge embedding weights $\gamma$ are optimized by alternating in the network and knowledge embedding spaces, respectively. At step $t$, given the current knowledge embedding $\gamma_{t-1}$, $w_t$ is optimized by descending the training loss $\mathcal{L}_{train}(w_{t-1}, \gamma_{t-1})$. Subsequently, $w_t$ is kept fixing and the over-parameterized knowledge embedding module is optimized with learning rate $\xi$ by descending 
\begin{align}
\nabla_{\gamma}\mathcal{L}_{val}(w_t-\xi\nabla_{w}\mathcal{L}_{train}(w_t,\gamma_{t-1}),\gamma_{t-1}).
\label{eq::val_loss}
\end{align} 
Finally, every over-parameterized knowledge embedding module is replaced as the knowledge embedding with the largest weight by taking argmax.

\subsection{Optimization Algorithm}
\label{sec::optimization}

Reinforcement learning \cite{pham2018efficient}, evolutionary algorithm \cite{shang2011novel,shang2015improved} and gradient-based \cite{liu2018darts} are three common optimization techniques for NAS. In this paper, we employ the combination of gradient-based optimization algorithm and the proposed Stacked BCNN to achieve extreme fast search efficiency. 

To discover a high-performance Stacked BCNN, the gradient-based optimization pipeline is constructed as: 1) a continuous relaxation strategy for over-parameterized Stacked BCNN construction, 2) partial channel connections (PC) for memory reduction and 3) early stopping for efficiency improvement. In the over-parameterized Stacked BCNN, a mixture operation with respect to all candidate operators is inserted between two nodes of each cell to construct the NAS pipeline for differentiable architecture search. In particular, we call the Stacked BCNN which consists of several mixture operation based mini BCNNs, as the over-parameterized Stacked BCNN. Moreover, the over-parameterized Stacked BCNN used in Stacked BNAS without KES (dubbed as case 1) and with KES (dubbed as case 2) are different. In case 1, the cell is over-parameterized, and the knowledge embedding is not over-parameterized as Fig. 5 (a). In case 2, for knowledge embedding search, both the cell and knowledge embedding are over-parameterized (see Fig. 5 (b)). The optimization algorithm of Stacked BNAS is shown in \textbf{Algorithm 1}.  

\subsubsection{Continuous Relaxation}

In mini BCNN, each cell consists of 2 input nodes $\{x_{(0)}, x_{(1)}\}$, $N-3$ intermediate nodes  $\{x_{(2)},\dots,x_{(N-2)}\}$ and a single output node $\{x_{(N-1)}\}$. Each intermediate node $x_{(i)}$ can be computed by
\begin{align}
x_{(i)} = \sum_{j<i}o_{(i,j)}(x_{(j)}),
\label{intermediate output}
\end{align}
where, $o_{(i,j)}$ is the operator between $x_{(i)}$ and $x_{(j)}$ chosen from candidate operation set $\mathcal{O}$. Moreover, the outputs of all intermediate nodes are combined to deliver $x_{(N-1)}$ by channel-dimension concatenation.

Subsequently, the over-parameterized Stacked BCNN is constructed by continuous relaxation \cite{liu2018darts}. Particularly, edge $(i,j)$ of each cell is relaxed for mini BCNN by
\begin{align}
f_{(i,j)}(x_{(j)}) = \sum_{o \in \mathcal{O}}\frac{{\rm exp}(\alpha_{(i,j)}^o)}{\sum_{o' \in \mathcal{O}}{\rm exp}(\alpha_{(i,j)}^{o'})}o(x_{(j)}),
\label{relaxation}
\end{align}
where operation $o(\cdot)$ is weighted by architecture weight $\alpha^o$.

\subsubsection{Partial Channel Connections}

The strategy of PC \cite{xu2019pc} is employed to improve the memory efficiency of Stacked BNAS and alleviate the performance collapse issue described in BNAS-v2 \cite{ding2021bnas-v2}. The continuous relaxation of Stacked BNAS with PC can be obtained by
\begin{align}
\begin{aligned}
f^{PC}_{(i,j)}(x_{(j)}; M_{(i,j)}) &= \sum_{o \in \mathcal{O}}\frac{{\rm exp}(\alpha_{(i,j)}^o)}{\sum_{o' \in \mathcal{O}}{\rm exp}(\alpha_{(i,j)}^{o'})}o(M_{(i,j)} * x_{(j)})\\
&+(1-M_{(i,j)})*x_{(j)},\\
\end{aligned}
\label{eq::pc}
\end{align}
where, $M_{(i,j)}$ represents the mask of the channel sample whose values are chosen from $\{0,1\}$. 

Moreover, edge normalization is used to mitigate the undesired fluctuation in the search phase via $\beta_{(i,j)}$ as
\begin{equation}
x_{(i)}^{PC}=\sum_{j<i}\frac{{\rm exp}(\beta_{(i,j)})}{\sum_{j'<i}{\rm exp}(\beta_{(i,j')})}\cdot f_{(i,j)}(x_{(j)}).
\label{xpc}
\end{equation}
Finally, each operation of the best architecture is obtained by taking argmax as
\begin{equation}
o_{(i,j)}=\mathop{{\rm argmax}}\limits_{o \in \mathcal{O}}\frac{{\rm exp}(\alpha_{(i,j)}^o)}{\sum_{o' \in \mathcal{O}}{\rm exp}(\alpha_{(i,j)}^{o'})} \cdot \frac{{\rm exp}(\beta_{(i,j)})}{\sum_{j'<i}{\rm exp}(\beta_{(i,j')})}.
\label{product}
\end{equation}

\subsubsection{Early Stopping (ES)}

As described in DARTS+ \cite{liang2019darts+}, there are two indices for ES: 1) the \textit{skip connection} number in a single cell and 2) the number of stable epochs. On the one hand, the search procedure is stopped when there is more than one \textit{skip connection} in one cell to avoid the performance collapse issue. PC contributes to reducing the predominance of \textit{skip connections}, so this paper does not choose the first index. On the other hand, the search procedure is stopped when the ranking of architecture weights is no longer changed for a determined number of epochs. This index means that the search procedure stops when arriving at a saturated state. Above all, the second index is chosen for early stopping of Stacked BNAS using the following criterion: 

\textbf{Criterion 1}: \textit{Stop the search procedure when the rank of architecture weights is no longer changed for three epochs.}    

Moreover, in the following sections, we employ repetitive time to represent the consistent epoch number where the rank of architecture weights is invariable.

\section{Universal Approximation of Stacked BCNN}
Given the initial input channel number $c$, the output of mini BCNN$_i$ with $C_i$ channels, i.e., \eqref{eq::mini_bcnn} can be rewritten as 
\begin{align}
y^{(i)}=\phi(\pmb{x}; \{\delta^{(i)}, \varphi^{(i)}, \pmb{W}^{(i)}_d, \pmb{\theta}^{(i)}_d, \pmb{W}^{(i)}_b, \pmb{\theta}^{(i)}_b,\pmb{W}^{(i)}_e, \pmb{\theta}^{(i)}_e \}), 
\label{eq::mini_bcnn_rewrite}
\end{align}
where $\pmb{x}$ represents input data. After GAP, each channel of $y^{(i)}$ is transformed into a single-pixel neuron-like feature map, so that it can be treated as $C_i$ neurons.   

Given standard hypercube $\textbf{I}^d = [0;1]^d \in \mathbb{R}^d$ and any continuous function $f \in {\rm C}(\textbf{I}^d)$, the proposed Stacked BCNN can be equivalently represented as
\begin{align}
f_{\pmb{p}_{k,u}}=\sum_{z=1}^{Z}w_z\sigma(\pmb{x}; \{\phi, \delta, \varphi, \pmb{W}^{(1)}, \pmb{\theta}^{(1)}, \dots, \pmb{W}^{(u)}, \pmb{\theta}^{(u)} \}),
\end{align}
where $Z=\sum_{i=1}^{u}C_i$ is the neuron number of the GAP output, $w$ represents the weight of the fully connected layer, $\pmb{p}_{k,u}=(k,u,c,w_1,\dots,w_{Z},\pmb{W}, \pmb{\theta})$ represents the set of overall parameters for the Stacked BCNN, and $\sigma$ is the activation function. Given the probability measure $\zeta_{k,u}$, randomly generates variables on $\pmb{\xi}_{k,u}=(w_1,\dots,w_{Z},\pmb{W}, \pmb{\theta})$. For compact set $\Omega$ of $\textbf{I}^d$, the distance between any continuous function and Stacked BCNN can be calculated as
\begin{align}
\chi_{\Omega}(f,f_{\pmb{p}_{k,u}})=\sqrt{\mathbb{E}\left[ \int_{\Omega}(f(\pmb{x})-f_{\pmb{p}_{k,u}}(\pmb{x}))^2d\pmb{x} \right]}.
\end{align}
Based on the above hypotheses, a theorem with proof of Stacked BCNN is given below.

\emph{Theorem 1}: Given any continuous function $f \in {\rm C}(\textbf{I}^d)$ and any compact set $\Omega \in \textbf{I}^d$, Stacked BCNN with nonconstant bounded functions $\phi,\delta,\varphi$, and absolutely integrable activation function $\sigma$ whose definition domain is $\textbf{I}^d$ so that $\int_{\mathbb{R}_d}\sigma^2(\pmb{x})d\pmb{x}< \infty$, has a sequence of $\{f_{\pmb{p}_{k,u}}\}$ with probability measures $\zeta_{k,u}$ satisfied that
\begin{align}
\mathop{\rm{lim}}\limits_{u \rightarrow \infty}\chi_{\Omega}(f,f_{\pmb{p}_{k,u}})=0.
\end{align}
Moreover, the trainable parameters $\pmb{\xi}_{k,u}$ are generated by $\zeta_{k,u}$.

\emph{Proof}: Define input data $\pmb{x}$, nonconstant bounded functions $\phi, \delta, \varphi$, approximation function $f_{\pmb{p}_{k,u'}}$ of Stacked BCNN with $u'$ mini BCNNs, probability distribution $\zeta_{k,u'}$ for trainable parameter generation, the weight matrix of fully connected layer $\pmb{w}'=[w'_{1},\dots,w'_{Z'}]^{\rm T}$ where $Z'=\sum_{z=1}^{u'}C_z$ and supplement weight $\pmb{w}''= [w''_{1},\dots,w''_{C_{u'+1}}]^{\rm T}$.

Stacked BCNN with $u'$ (any integer) mini BCNNs can be computed by
\begin{align}
f_{\pmb{w}'}=&\sum_{z=1}^{Z'}w'_{z}\sigma(\pmb{x}; \{\phi, \delta, \varphi, \pmb{W}^{(1)}, \pmb{\theta}^{(1)}, \dots, \pmb{W}^{(u')}, \pmb{\theta}^{(u')} \}).
\end{align}
Subsequently, Stacked BCNN with input data $\pmb{x}$ can approximate continuous function $f$ with bounded and integrable resident function $f_{r_{u'}} \in \textbf{I}^d$ as
\begin{align}
f_{r_{u'}}(\pmb{x}) = f(\pmb{x})-f_{\pmb{w}'}(\pmb{x}).
\end{align}
As described in previous work \cite{rudin2006real}, for $\forall \varepsilon>0$, a function $f_{b_{u'}} \in {\rm C}(\textbf{I}^d)$ can always be found to satisfy:
\begin{align}
\chi_{\Omega}(f_{b_{u'}},f_{r_{u'}})<\frac{\varepsilon}{2}.
\label{buru}
\end{align}

An extra mini BCNN (i.e., mini BCNN$_{u'+1}$) is defined to approximate $f_{b_{u'}}$ with $C_{u'+1}$ channels. Mini BCNN$_{u'+1}$ can be equivalently expressed as
\begin{align}
f_{\pmb{w}''}=\sum_{z=1}^{C_{u'+1}}w''_{z}\underbrace{\sigma(\pmb{x}; \{\phi, \delta, \varphi, \pmb{W}^{(u'+1)}, \pmb{\theta}^{(u'+1)} \})}_{\vartheta},
\label{fwe}
\end{align}
Similarly, the composition function $\vartheta$ in \eqref{fwe} is absolutely integrable. According to \emph{Theorem 1} in \cite{igelnik1995stochastic}, for $\forall \varepsilon>0$, a sequence of $f_{\pmb{w}''}$ can be found to satisfy:
\begin{align}
\chi_{\Omega}(f_{b_{u'}},f_{\pmb{w}''})<\frac{\varepsilon}{2}.
\end{align}
Moreover, the output of Stacked BCNN can be rewritten as
\begin{align}
f_{\pmb{p}_{k,u}} = f_{\pmb{w}'}+f_{\pmb{w}''}.
\end{align}

The distance between $f$ and  $f_{\pmb{p}_{k,u}}$ can be obtained by
\begin{align}
\begin{aligned}
\chi_{\textbf{I}^d}(f,f_{\pmb{p}_{k,u}})&=\sqrt{\mathbb{E}\left[ \int_{\Omega}(f(\pmb{x})-f_{\pmb{p}_{k,u}}(\pmb{x}))^2d\pmb{x} \right]}\\
&= \sqrt{\mathbb{E}\left[ \int_{\Omega}\left((f(\pmb{x})-f_{\pmb{w}'}(\pmb{x}))-f_{\pmb{w}''}(\pmb{x})\right)^2d\pmb{x} \right]}\\
&= \sqrt{\mathbb{E}\left[ \int_{\Omega}\left(f_{r_{u'}}-f_{\pmb{w}''}\right)^2d\pmb{x} \right]}\\
&= \chi_{\Omega}(f_{r_{u'}},f_{\pmb{w}''})\\
&\leq \chi_{\Omega}(f_{b_{u'}},f_{r_{u'}}) + \chi_{\Omega}(f_{b_{u'}},f_{\pmb{w}''}) \\
&< \frac{\varepsilon}{2} + \frac{\varepsilon}{2} \\
&<\varepsilon\\
\end{aligned}
\end{align}
Therefore, a conclusion can be drawn as
\begin{align}
\mathop{\rm{lim}}\limits_{u,v\rightarrow \infty}\chi_{\Omega}(f,f_{\pmb{p}_{k,u}})=0.
\end{align}
In other words, the proposed Stacked BCNN can completely appropriate any continuous function.

\section{Experiments and Results}
\label{sec::Experiments}

In this section, the datasets used and implementation details are given first. Next, architecture search of Stacked BNAS without/with KES are introduced. Then, the best-performing architecture learned by Stacked BNAS on CIFAR-10 is transferred to solve large-scale image classification task on ImageNet, and the experimental results are analysed. Furthermore, the learned architecture is also transferred to four image classification datasets to verify the generalization ability of the proposed Stacked BNAS. Finally, two groups of ablation experiments are performed to verify the effectiveness of Stacked BNAS in terms of efficiency and performance. 

\subsection{Datasets and Implementation Details}
\subsubsection{Datasets}

CIFAR-10 and ImageNet are used to verify the effectiveness of the proposed Stacked BNAS. CIFAR-10 is a small-scale image classification dataset with 32$\times$32 pixels that contains 50K training images and 10K test images. ImageNet contains approximately 1.3 M training data and 50K validation data with various pixels over 1000 object categories. 

\subsubsection{Implementation Details}

The data preprocessing technique follows BNAS-v2 for CIFAR-10 and ImageNet. Architecture search without or with KES ar e performed. For architecture search, the implementation is repeated with five times. For architecture evaluation, the mean value of three repetitive retraining experiments is treated as the index to determine the best architecture. Furthermore, the best architecture learned on CIFAR-10 is transferred to ImageNet, MNIST, FashionMNIST, NORB and SVHN. 

\subsection{Experiments on CIFAR-10}

\subsubsection{Experimental Settings}

As mentioned above, two experiments are implemented on CIFAR-10: 1) Stacked BNAS without KES and 2) Stacked BNAS with KES. 

The above two experiments use many identical experimental settings for architecture search, as shown below. The initial input channel number is set to 16. The above over-parameterized Stacked BCNN is trained for 50 epochs. All training images are equally divided into two parts. On the one hand, 25K images are treated as the training data to update the network weights $w$. On the other hand, another part is used as the validation data to optimize the architecture/embedding weights. To optimize the network weights $w$, the SGD optimizer is used with a dynamic learning rate using an annealed decay manner, momentum of 0.9 and weight decay of 3$\times$10$^{-4}$. Beyond that, the Adam optimizer is used to update the architecture/embedding weights, with weight decay of 1$\times$10$^{-3}$.  Architecture search is implemented on a single  NVIDIA GTX 1080Ti GPU. 

For architecture search of Stacked BNAS without KES, the over-parameterized Stacked BCNN consists of 2 mini BCNNs, where each one contains 1 broad cell and 1 enhancement cell. The batch size and network learning rate are set to 512 and 0.2, respectively. Moreover, the architecture learning rate is set to 6$\times$10$^{-4}$. For architecture search of Stacked BNAS with KES, the over-parameterized Stacked BCNN consists of 2 mini BCNNs, where each one contains 1 deep cell, 1 broad cell and 1 enhancement cell. Due to more memory requirements of KES, the batch size and network learning rate are set to 128 and 0.05, respectively. Beyond that, a larger learning rate of 2$\times$10$^{-3}$ is set for architecture/embedding weights when knowledge embedding is learned. 

\begin{figure}[!t]
  \centering
    \subfloat[Convolution cell]{\includegraphics[width=0.47\textwidth]{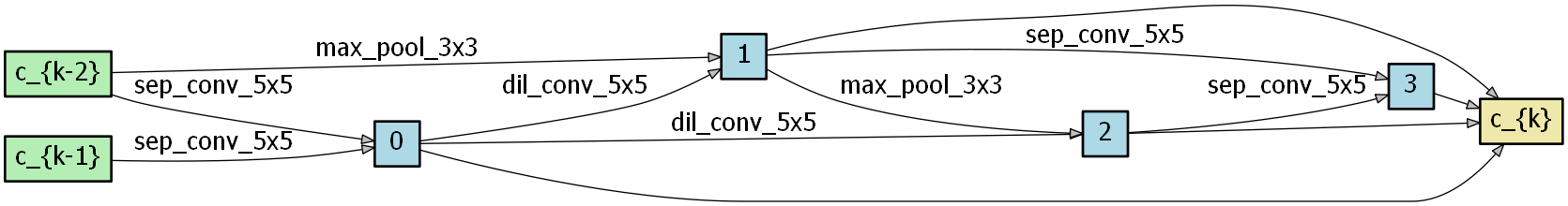}} \\
    \subfloat[Enhancement cell]{\includegraphics[width=0.47\textwidth]{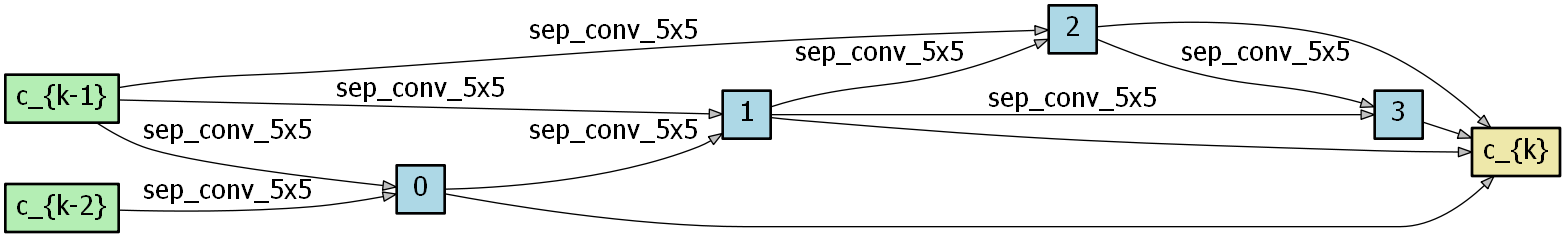}} \\
\captionsetup{font={small}}   
  \caption{Architecture learned by Stacked BNAS without KES.}
    \label{fig::cells}
     \vspace{-0.5cm}
\end{figure}

\begin{figure}[!t]
  \centering
    \subfloat[Convolution cell]{\includegraphics[width=0.27\textwidth]{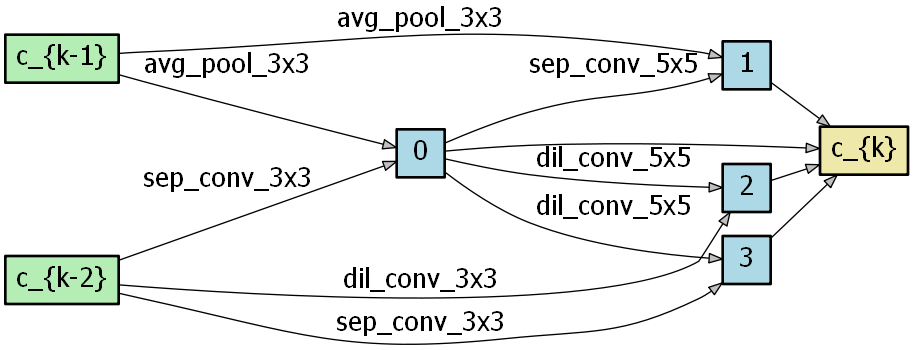}} \\
    \subfloat[Enhancement cell]{\includegraphics[width=0.37\textwidth]{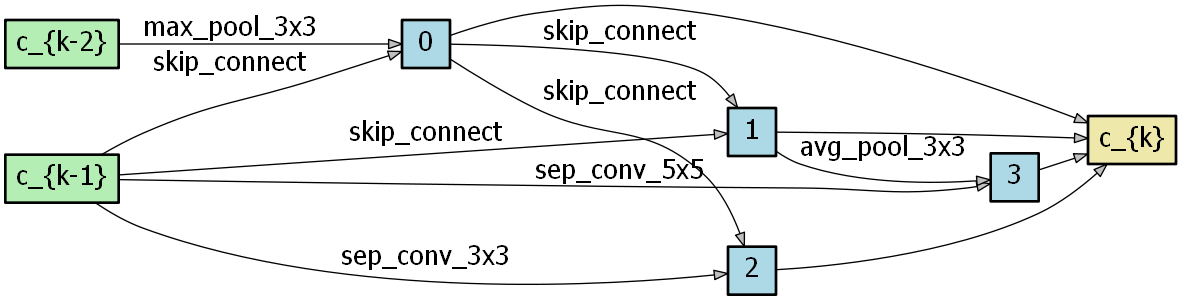}} \\
\captionsetup{font={small}}   
  \caption{Architecture learned by Stacked BNAS with KES.}
    \label{fig::cells_embedding}
     \vspace{-0.5cm}
\end{figure}

\begin{table*}[!t]
\centering
\small
\begin{threeparttable}[tbq]
\captionsetup{font={small}} 
\caption{Comparison of the proposed Stacked BNAS and other state-of-the-art NAS approaches on CIFAR-10.}
\label{tab::cifar10}
\begin{tabular}{lcccccc}
\hline
\multirow{2}{*}{\textbf{Architecture}} & \textbf{Error}& \textbf{Params}& \textbf{Search Cost} &\textbf{Number} & \multirow{2}{*}{\textbf{Search Method}} & \multirow{2}{*}{\textbf{Topology}}\\
& \textbf{(\%)} & \textbf{(M)}   & \textbf{(GPU days)} & \textbf{of Cells} & \\
\hline
LEMONADE \cite{elsken2018efficient}    & 3.05 & 4.7  & 80  & - & evolution  & deep\\
DARTS (1st order) \cite{liu2018darts} & 3.00 & 3.3  & 0.45$\dag$ & 20 & gradient-based & deep\\
DARTS (2nd order) \cite{liu2018darts} & 2.76 & 3.3  & 1.50$\dag$ & 20 & gradient-based & deep\\
DARTS (random) \cite{liu2018darts}& 3.49 & 3.1  & - & 20 & - & deep\\
SNAS + mild constraint \cite{xie2018snas}  & 2.98 & 2.9  & 1.50  & 20 & gradient-based  & deep\\
SNAS + moderate constraint \cite{xie2018snas}  & 2.85 & 2.8  & 1.50 & 20 &  gradient-based & deep\\
SNAS + aggressive constraint \cite{xie2018snas}  & 3.10 & \textbf{2.3}  & 1.50 & 20 &  gradient-based & deep\\
P-DARTS \cite{chen2019progressive} & \textbf{2.50} & 3.4  & 0.30 & 20  & gradient-based& deep\\
GDAS-NSAS \cite{zhang2020overcoming} & 2.73 & 3.5 & 0.40 & 20 & gradient-based & deep \\
PC-DARTS \cite{xu2019pc} & 2.57 & 3.6  & 0.10 & 20 &  gradient-based & deep\\
ENAS \cite{pham2018efficient} & 2.89 & 4.6  & 0.45 & 17 & RL & deep\\
\hline
\hline
BNAS \cite{ding2021bnas}      & 2.97 & 4.7 & 0.20  & \textbf{5} & RL & broad\\
BNAS-CCLE \cite{ding2021bnas} & 2.95 & 4.1 & 0.20 & \textbf{5} &  RL & broad\\
BNAS-CCE \cite{ding2021bnas}  & 2.88 & 4.8 & 0.19 & 8 & RL & broad\\
BNAS-v2 \cite{ding2021bnas-v2} & 2.79 & 3.7 & 0.05 & 8 & gradient-based   & broad\\
\hline
\hline
Random  & 3.12 & 3.1 & - & 8 & - & broad\\
Stacked BNAS w/o KES (Ours)  & 2.71 & 3.7 & \textbf{0.02} & 8 & gradient-based & broad\\
Stacked BNAS w/ KES (Ours)  & 2.73 & 3.1 & 0.15 &  10 &gradient-based & broad\\
\hline
\end{tabular}
\footnotesize
\begin{tablenotes}
\item[$\dag$] Obtained by DARTS using the code publicly released by the authors at https://github.com/quark0/darts on a single NVIDIA GTX 1080Ti GPU.
\end{tablenotes}
\end{threeparttable}
\end{table*}

For architecture evaluation, identical settings are employed except the number of deep cell which is 2 and 3 for Stacked BNAS without/with KES, respectively.  The stacked BCNN consists of two mini BCNNs, where each one contains 1 broad cell and 1 enhancement cell. Following BNAS-v2, the learned Stacked BCNN with 44 input channels is trained for 2000 epochs using the SGD optimizer with a batch size of 128, learning rate of 0.05, momentum of 0.9 and weight decay of 3$\times$10$^{-4}$.  Moreover, architecture evaluation is implemented on a single  NVIDIA Tesla P100 GPU.

\subsubsection{Results and Analysis}

\begin{figure}[!t]
  \centering
    \subfloat[The first mini BCNN]{\includegraphics[width=0.43\textwidth]{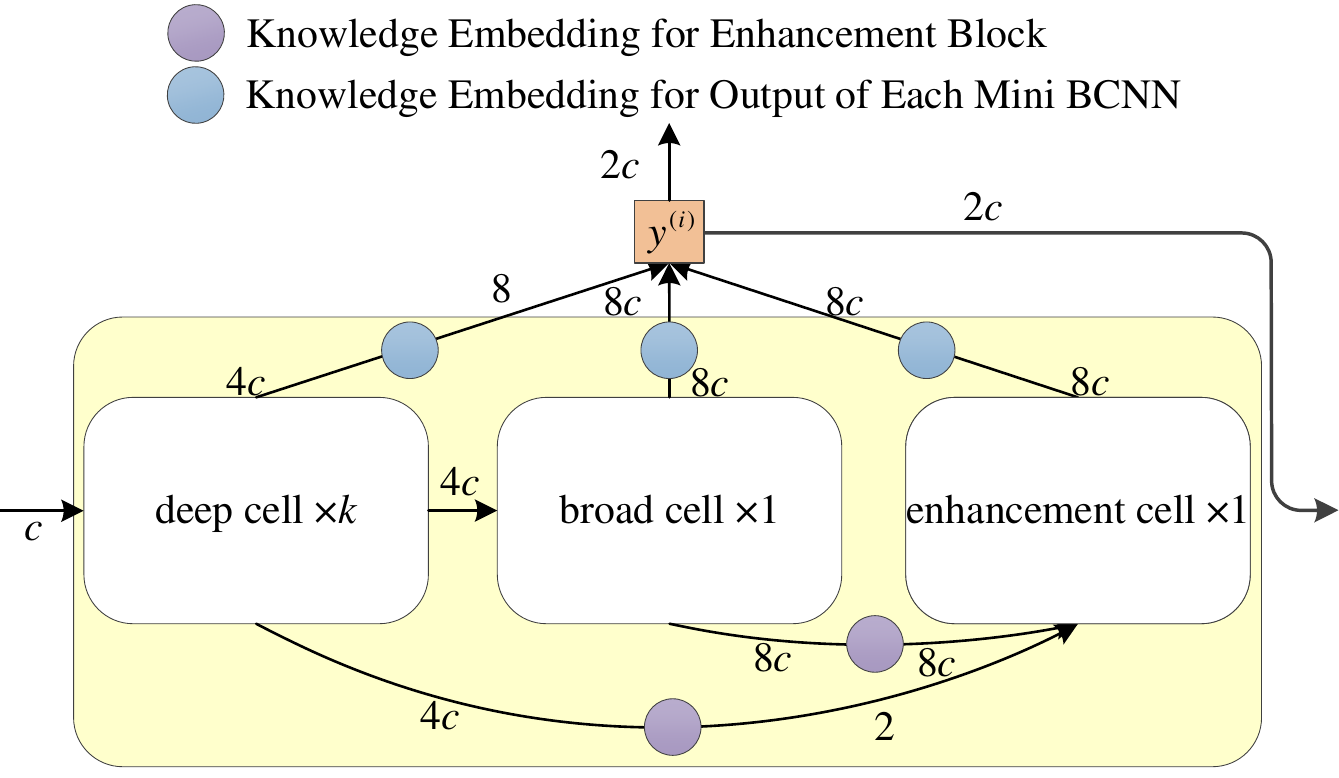}} \\
    \subfloat[The second mini BCNN]{\includegraphics[width=0.43\textwidth]{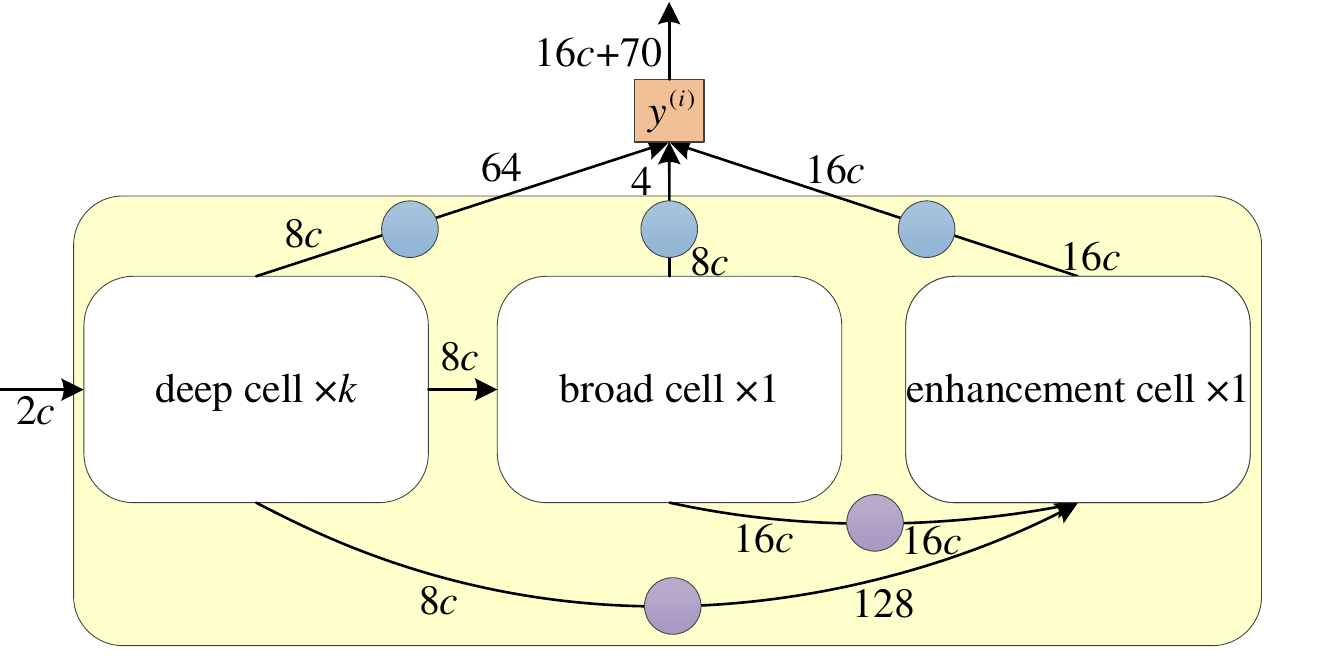}} \\
  \caption{The knowledge embedding learned by Stacked BNAS with KES, where $c=44$.}
    \label{fig::embedding}
     \vspace{-0.5cm}
\end{figure}

For Stacked BNAS, the learned architecture is visualized in Fig. \ref{fig::cells}. For Stacked BNAS with KES, the best architecture and knowledge embedding are shown in Fig. \ref{fig::cells_embedding} and \ref{fig::embedding}, respectively. Furthermore, TABLE \ref{tab::cifar10} summarizes the comparison of the proposed Stacked BNAS with other novel NAS approaches on CIFAR-10. 

Contributing to the combination of Stacked BCNN and optimization strategy, Stacked BNAS delivers a state-of-the-art efficiency of 0.02 GPU days and competitive test accuracy of 97.29\% with 3.7 M parameters. Beyond that, Stacked BNAS with KES discovers a Stacked BCNN with 2.73\% test error and just 3.1 M parameters (approximately 16.2\% less than Stacked BNAS). Moreover, the over-parameterized knowledge embedding module leads to more trainable parameters and memory requirements than vanilla Stacked BNAS, so larger costs are needed. As shown in Fig. \ref{fig::embedding}, one indirect-connected knowledge embedding of each mini BCNN has more output channels than hand-crafted embedding. The above knowledge embedding changes lead to parameter reduction of the architecture learned by Stacked BNAS with KES.  

Compared with NAS approaches with deep topology, Stacked BNAS obtains the best efficiency and competitive accuracy. In terms of random architecture, Stacked BCNN obtains 0.38\% better accuracy than DARTS, which further examines the effectiveness of the proposed Stacked BCNN. Furthermore, Stacked BNAS is 5$\times$ faster than PC-DARTS whose efficiency ranks the best among all NAS approaches. Compared with BNASs, the proposed Stacked BNAS delivers better efficiency and accuracy. On the one hand, the efficiency of Stacked BNAS is approximately 10$\times$ and 2.5$\times$ faster than BNAS-v1 and BNAS-v2, respectively. On the other hand, the accuracy of Stacked BNAS is approximately 0.2\% and 0.1\% better than BNAS-v1 and BNAS-v2, respectively. Compared with previous BNASs, Stacked BNAS with KES can deliver better accuracy with approximately 16\% and 33\% parameter reduction, respectively.

\subsubsection{Efficiency Difference between Stacked BNAS without or with KES}

As shown in TABLE \ref{tab::cifar10}, the efficiencies of Stacked BNAS without/with KES are 0.02 and 0.15 GPU days, respectively. For that, there are two main reasons: 1) different structures of used Stacked BCNNs in the search phase, and 2) various degrees of difficulty meeting \textbf{Criterion 1}. 

On the one hand, in the architecture search phase of Stacked BNAS with KES, each mini BCNN contains a deep cell for learning appropriate knowledge embedding. Consequently, the first advantage of fast single-step training speed does not work. On the other hand, Stacked BNAS with KES has difficult satisfying \textbf{Criterion 1} for early stopping.

\begin{table*}[!t]
\centering
\small
\begin{threeparttable}[tbq]
\captionsetup{font={small}} 
\caption{Comparison of the proposed Stacked BNAS and other state-of-the-art NAS approaches on ImageNet.}
\label{tab::imagenet}
\begin{tabular}{lccccccc}
\hline
\multirow{2}{*}{\textbf{Architecture}} & \multicolumn{2}{c}{\textbf{Test Err. (\%)}} & \textbf{Params} & \textbf{Search Cost} & \textbf{FLOPs} & \multirow{2}{*}{\textbf{Search Method}} & \multirow{2}{*}{\textbf{Topology}} \\
\cline{2-3}
& \textbf{top-1} & \textbf{top-5} & \textbf{(M)}  & \textbf{(GPU days)} & \textbf{(M)} \\
\hline
AmoebaNet-A \citep{real2018regularized}& 25.5 & 8.0 & 5.1 & 3150 & 555 & evolution & deep\\
NASNet-A \citep{zoph2018learning}      & 26.0 & 8.4 & 5.3 & 1800 & 564 & RL & deep\\
PNAS \citep{liu2018progressive}        & 25.8 & 8.1 & 5.1 & 225  & 588 & SMBO & deep\\
GHN \citep{zhang2019graph}        & 27.0 & 8.7 & 6.1 & 0.84  & 569 & SMBO & deep\\
DARTS (2nd order) \citep{liu2018darts}         & 26.7 & 8.7 & 4.7 & 1.50  & 574 & gradient-based & deep\\
SNAS (mild) \citep{xie2018snas}                & 27.3 & 9.2 & 4.3 & 1.50  & 522 & gradient-based & deep \\
BayesNAS \citep{zhou2019bayesnas} & 26.5 &  8.9  & \textbf{3.9} & 0.20 & - & gradient-based & deep\\
PC-DARTS \citep{xu2019pc}            & 25.1 &  7.8  & 5.3 & 0.10 & 586 & gradient-based & deep\\
PC-DARTS (ImageNet) \citep{xu2019pc}     & \textbf{24.2} &  \textbf{7.3}  & 5.3 & 3.80 & 597 & gradient-based & deep\\
BNAS-v2 (PC) (2nd order)  \cite{ding2021bnas-v2}     & 27.2 & 8.8 & 4.6 & 0.09 & \textbf{475} & gradient-based & broad \\
\hline
\hline
Stacked BNAS (Ours)     & 26.4 & 8.9 & 4.7 & \textbf{0.02} & 485 & gradient-based & broad \\
\hline
\end{tabular}
\end{threeparttable}
\end{table*}

\begin{figure}[!t]
\centering
\captionsetup{font={small}} 
\includegraphics[width=0.41\textwidth]{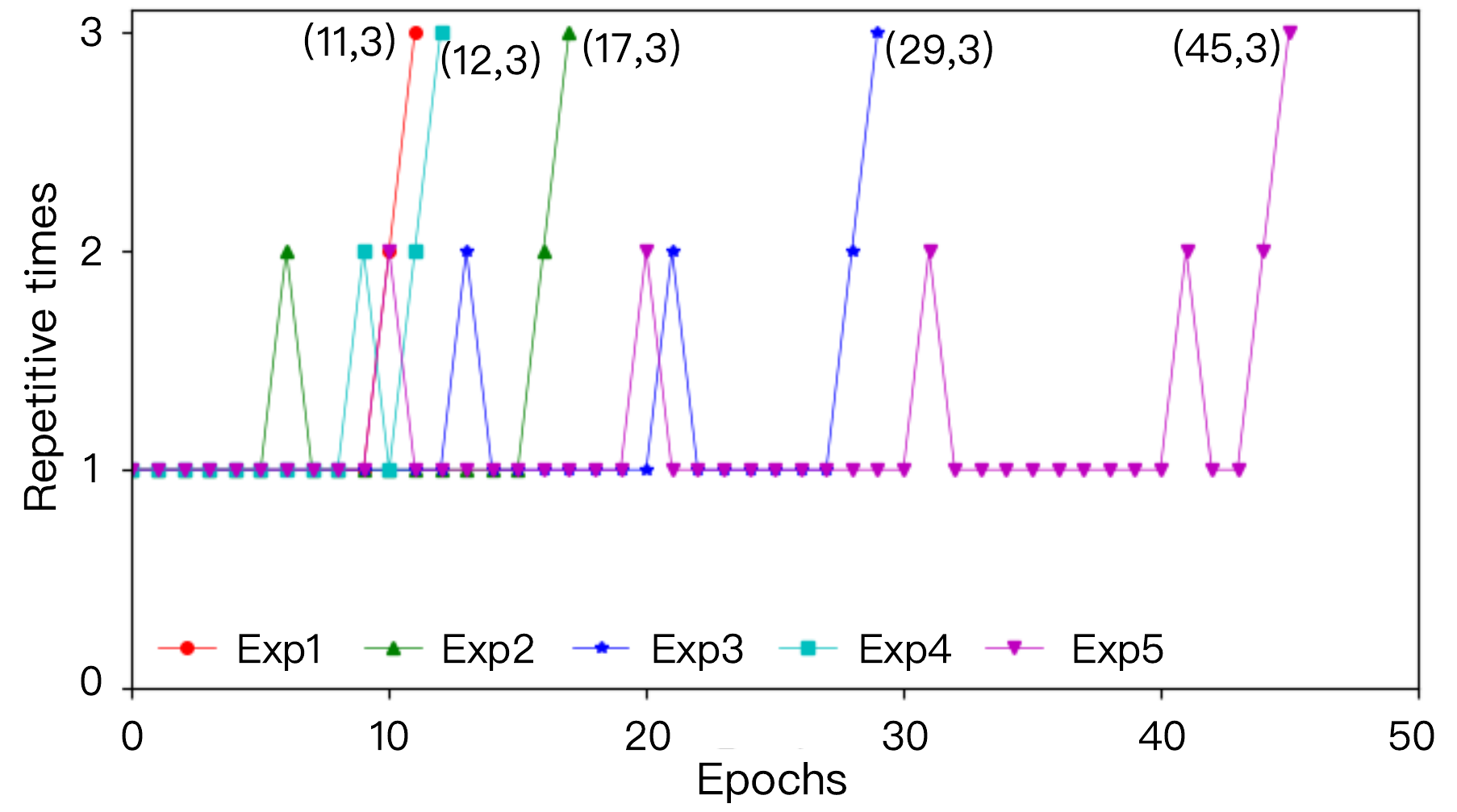}
\caption{Early stopping for Stacked BNAS without KES. For instant, (12, 3) means that the proposed approach can satisfy the early stopping condition of 3 epochs when arriving at 12-th epoch.}
\label{fig::ES}
 \vspace{-0.5cm}
\end{figure} 

\begin{figure}[!t]
  \centering
  \small
    \subfloat[Repetitive times of both architecture and embedding]{\includegraphics[width=0.41\textwidth]{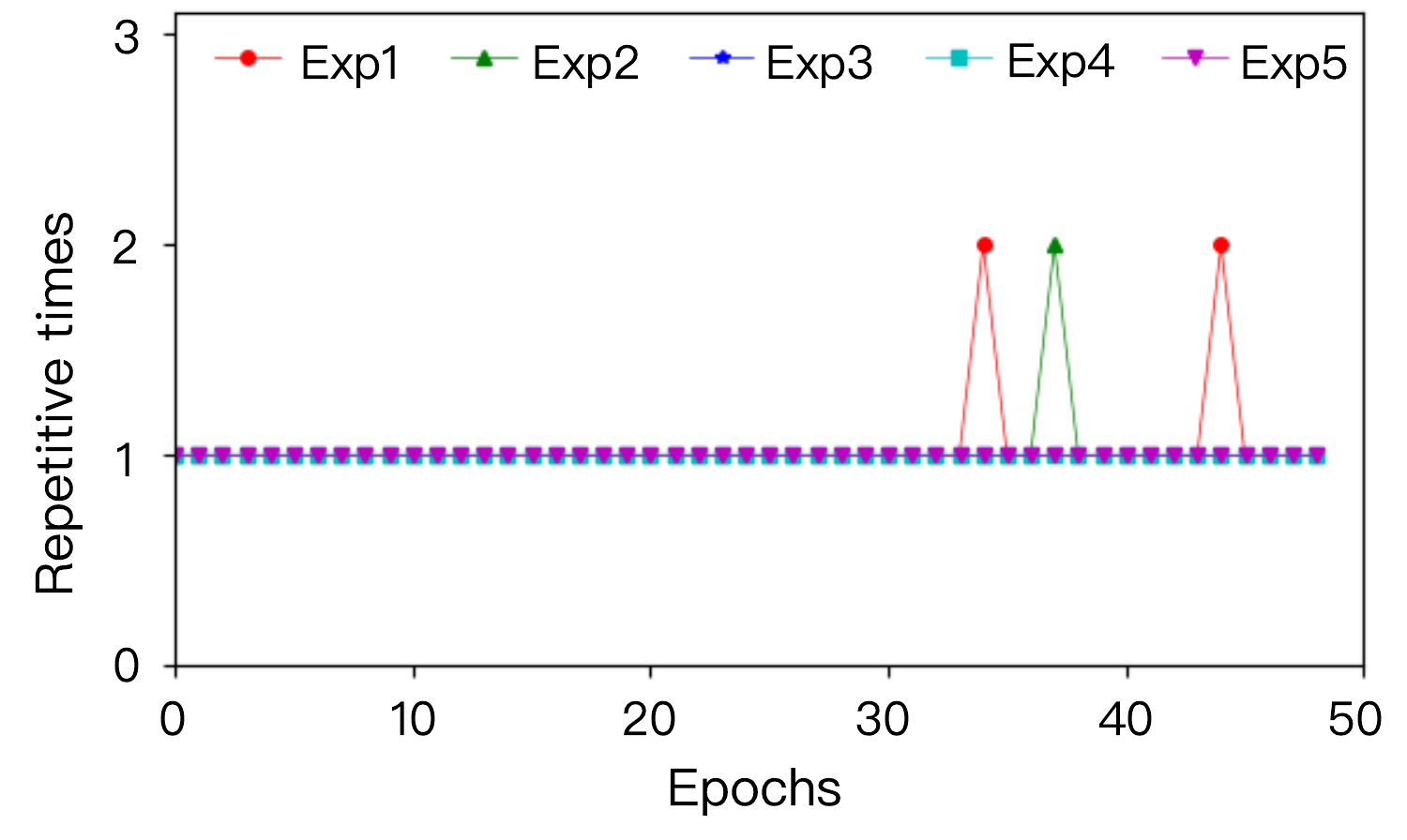}} \\
    \subfloat[Repetitive times of architecture]{\includegraphics[width=0.41\textwidth]{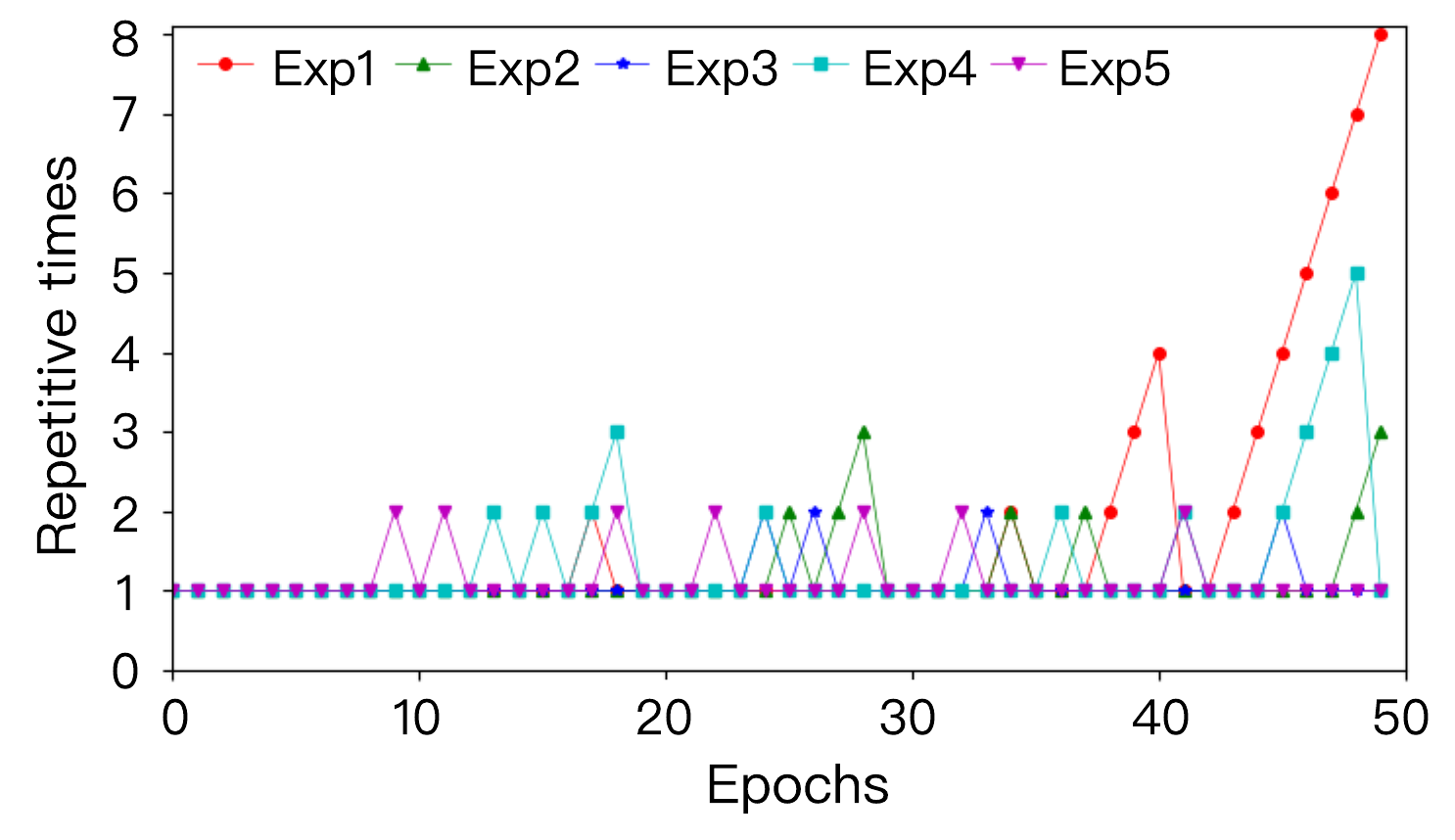}} \\
    \subfloat[Repetitive times of embedding]{\includegraphics[width=0.41\textwidth]{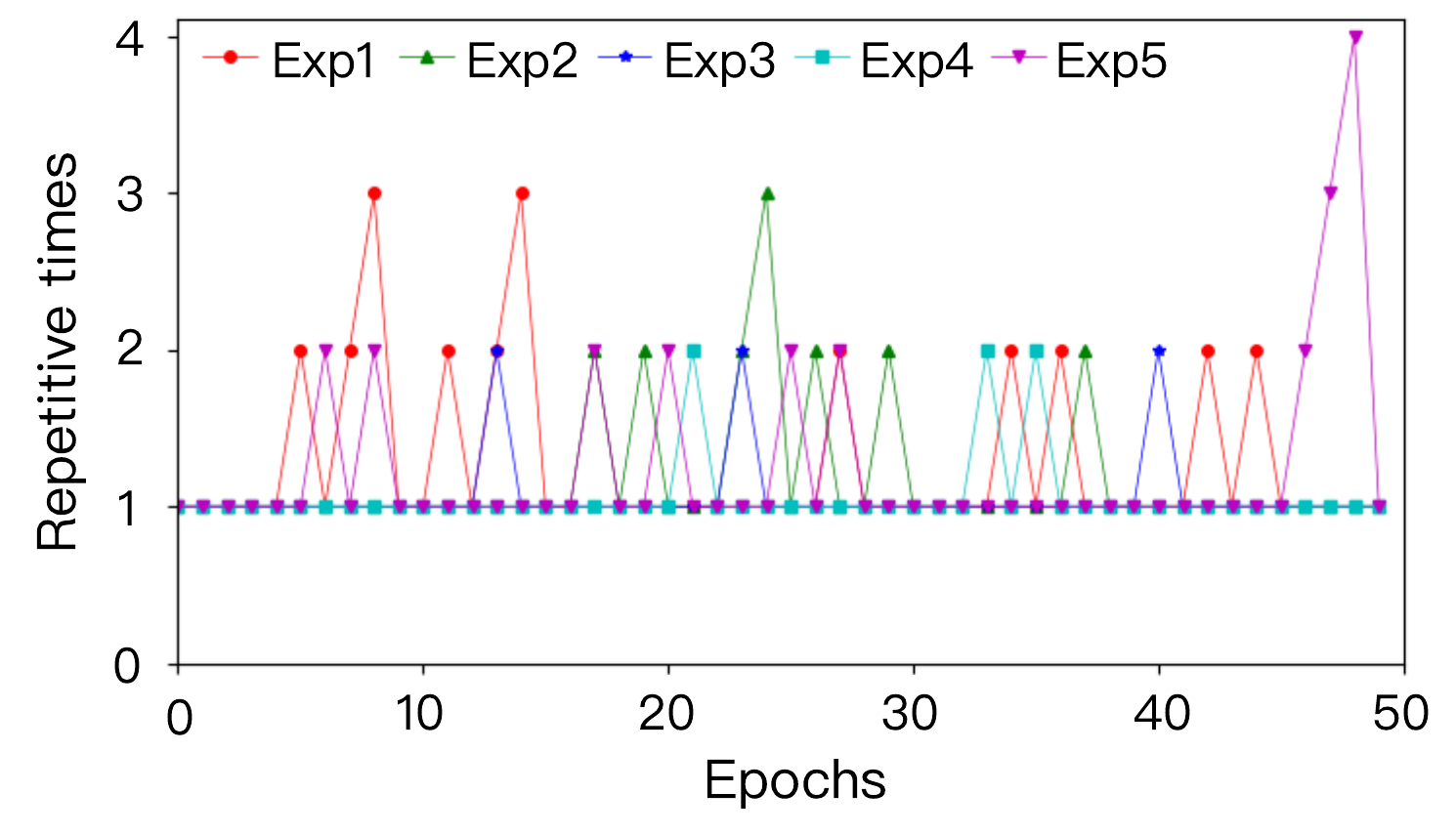}} \\  
\captionsetup{font={small}}   
  \caption{Early stopping for Stacked BNAS with KES where both architecture and embedding weight should satisfy the \textbf{Criterion 1} \textit{simultaneously}.}
     \vspace{-0.5cm}
    \label{fig::ES_kes}
\end{figure}

\begin{itemize}
\item The architecture repetitive times of Stacked BNAS are shown in Fig. \ref{fig::ES}. In the architecture search phase of Stacked BNAS, both the repetitive times of convolution and enhancement cells should be larger than 2. Each implementation can stop early before arriving at the maximum epoch so that Stacked BNAS delivers state-of-the-art efficiency. 
\item For Stacked BNAS with KES, the architecture and embedding repetitive times are shown in Fig. \ref{fig::ES_kes}. In this experiment, the ES strategy is that the convolution cell, enhancement cell,  and knowledge embedding do not change in three epochs. Each implementation cannot satisfy the aforementioned early stopping criterion. Beyond that, over-parameterized knowledge embedding without a partial channel connections strategy leads to more memory usages than vanilla BNAS. As a result, the batch size is set as 128 instead of 256. Above all, the efficiency of Stacked BNAS is not satisfactory when using KES.
\end{itemize}     

\subsection{Experiments on ImageNet}

\subsubsection{Experimental Settings}

To transfer the architecture learned by Stacked BNAS on ImageNet, three 3$\times$3 convolutions are treated as the stem layers that reduce the input size from 224$\times$224 to 28$\times$28. Subsequently, a Stacked BCNN is constructed by 2 mini BCNNs, where each one contains 2 deep cells, 1 broad cell and 1 enhancement cell. 

For architecture evaluation, the SGD optimizer is chosen with a learning rate of 0.1 followed by a linear decay method, momentum of 0.9 and weight decay of 3$\times$10$^{-5}$. Moreover, the Stacked BCNN is trained for 250 epochs with a batch size of 512 on 4 NVIDIA Tesla P100 GPUs. Other experimental settings follow PC-DARTS. 

\subsubsection{Results and Analysis}


TABLE \ref{tab::imagenet} summarizes the comparison of the proposed Stacked BNAS with other novel NAS approaches on ImageNet. 

Compared with previous impactful NAS approaches \cite{zoph2018learning,real2018regularized,liu2018progressive}, the proposed Stacked BNAS delivers competitive or better performance with a state-of-the-art efficiency of 0.02 GPU days which is 5 or 6 magnitudes faster. For efficient NAS approaches \cite{liu2018darts,xu2019pc}, Stacked BNAS also obtains competitive or better performance with 1 or 2 magnitudes faster efficiency. Compared with BNAS-v2, Stacked BNAS obtains better performance in terms of top-1 and top-5 accuracy. As mentioned above, the main difference between BNAS-v2 and the proposed Stacked BNAS is the broad scalable architecture, so that the effectiveness of Stacked BCNN can be promised. Moreover, the search cost of BNAS-v2 is 4.5$\times$ higher than Stacked BNAS. Due to the broad topology of Stacked BCNN, Stacked BNAS delivers the best performance in terms of FLOPs, which is an important index to show the  hardware friendliness of deep models.

\subsection{Experiments on Other Datasets}

To further verify the effectiveness of Stacked BNAS, similar with BNAS-v2, the architecture learned on CIFAR-10 is transferred to MNIST, FashionMNIST, NORB and SVHN. Stacked BNAS employs identical experimental settings for the above four datasets. The Stacked BCNN consists of 2 mini BCNNs where each one has 3 deep cells, 1 broad cells and 1 enhancement cells. SGD is chosen as the optimizer to train the Stacked BCNN from scratch for 300 epochs. Here, several important hyper-parameters are listed as follows: a batch size of 128 and an initial learning rate of 0.05. Other training hyper-parameters are identical to BNAS-v2. Similar to BNAS-v2, deep search space-based NAS approaches (e.g., NASNet \cite{zoph2018learning}, AmoebaNet \cite{real2018regularized}, DARTS \cite{liu2018darts} and PC-DARTS \cite{xu2019pc}) consist of 20 cells and employ identical settings to Stacked BNAS except the learning rate of 0.025. Moreover, BNAS-v2 consists of 8 cells and sets the learning rate to 0.025. Experimental results are shown in TABLE \ref{tab::datasets}.

\begin{table*}[!t]
\centering
\small
\begin{threeparttable}[tbq]
\captionsetup{font={small}} 
\caption{Comparison of Stacked BNAS and other novel classifiers on four image classification datasets}
\label{tab::datasets}
\begin{tabular}{lccccccc}
\hline
\multirow{2}{*}{\textbf{Architecture}} &  \textbf{Params} & \multicolumn{4}{c}{\textbf{Accuracy (\%)}} & \textbf{Number} & \textbf{Search Cost} \\
\cline{3-6}
 & \textbf{(M)} & \textbf{MNIST} & \textbf{FashionMNIST} & \textbf{NORB} & \textbf{SVHN} & \textbf{of Cells} & \textbf{(GPU Days)}\\
\hline
NASNet \cite{zoph2018learning}   & 1.5/1.3$\dag$ & \textbf{99.64} (1) & \textbf{95.47} (1) & 93.34 (4) & 96.87 (4) & 20 & 1800\\
AmoebaNet \cite{real2018regularized}  & 1.5 & 99.62 (4) & 95.33 (3) & \textbf{93.73} (1) & 96.85 (5) & 20 & 3150\\
DARTS \cite{liu2018darts} & 1.5 & 99.58 (6) & 95.24 (6) & 91.83 (6) & 96.76 (6) & 20 & 0.45\\
PC-DARTS \cite{xu2019pc} & 1.4 & 99.61 (5) & 95.26 (5) & 93.00 (5) & 96.98 (2) & 20 & 0.1\\
BNAS-v2 \cite{ding2021bnas-v2} & 1.5 & 99.63 (3) & 95.33 (3) & 93.37 (3) & 96.98 (2) & \textbf{8} & 0.09\\
\hline
Stacked BNAS (ours) & 1.5 & \textbf{99.64} (1) & 95.35 (2) & 93.52 (2) & \textbf{97.12} (1) & 10 & \textbf{0.02} \\
\hline
\end{tabular}
\footnotesize
\begin{tablenotes}
\item[$\dag$] The error of out of memory arises on a NVIDIA Tesla P100 GPU when using 1.5 M parameters for NORB and SVHN classification.
\end{tablenotes}
\end{threeparttable}
\end{table*}

Compared with the comparative approaches, Stacked BNAS delivers the best generalization ability. For MNIST, both Stacked BNAS and NASNet obtain the best performance, i.e., 99.64\% accuracy. For FashionMNIST, NASNet also delivers the highest accuracy. The accuracy of Stacked BNAS is second-ranked. For NORB, the accuracy of Stacked BNAS is also second-ranked and AmoebaNet performs the best. For SVHN, Stacked BNAS is tied for the first place with 97.12\% accuracy.  Above all, Stacked BNAS shows the best generalization ability on all tasks.

\subsection{Ablation Studies}

Here, two groups of experiments are implemented: 1) one is to examine the effectiveness of PC and ES for efficiency improvement of Stacked BNAS, and 2) the other is to verify the  effectiveness of two scales of information, i.e., the output of the broad cell to the output node is denoted as b2o and the output of the deep cell to the enhancement cell is denoted as d2e. 

\subsubsection{PC and ES for Efficiency Improvement}

In this group of experiments, there are four cases: 1) using neither PC nor ES, 2) using only PC, 3) using only ES and 4) using both PC and ES. Each case is repeatedly performed five times following the experimental setting used for architecture search of Stacked BNAS without KES. Experimental results are shown in TABLE \ref{tab:efficiency_ablation}. 

\begin{table}[!h]
\centering
\small
\captionsetup{font={small}} 
\caption{Ablation experiments for the efficiency of Stacked BNAS on CIFAR-10.}
\label{tab:efficiency_ablation}
\begin{tabular}{cccccc}
\hline
\multirow{2}{*}{\textbf{Case}} & \multirow{2}{*}{\textbf{PC}} & \multirow{2}{*}{\textbf{ES}} & \multirow{2}{*}{\textbf{Epochs}} & \multirow{2}{*}{\textbf{Batch Size}} & \textbf{Efficiency}\\
&&&&&\textbf{(GPU days)}\\
\hline
\hline
1 & \XSolidBrush    & \XSolidBrush  &  50 &  128  &  0.140 \\
2 & \Checkmark   &  \XSolidBrush  & 50  & \textbf{512} & 0.068 \\
3 & \XSolidBrush     & \Checkmark  &  15 &  128  &  0.047 \\
4 & \Checkmark  & \Checkmark & \textbf{12}  & \textbf{512} & \textbf{0.018} \\
\hline
\end{tabular}
\end{table}

In case 1, the search cost of Stacked BNAS without PC and ES is 0.14 GPU days under a single NVIDIA GTX 1080Ti GPU. In case 2, Stacked BNAS employs PC to deliver an efficiency of 0.068 GPU days, which is 2$\times$ faster than case 1, because the strategy of PC contributes to improving a higher memory efficiency (using a larger batch size, i.e., 512) of Stacked BNAS than case 1. In the case of using ES, Stacked BNAS can stop the search phase at the 15th epoch and obtain an efficiency of 0.047 GPU days. When using both PC and ES, Stacked BNAS can not only search architecture with efficient memory of 512 batch size, but also stop early at the 12th epoch and delivers a state-of-the-art efficiency of 0.018 GPU days. Above all, both PC and ES play important roles in the efficiency improvement of Stacked BNAS.

\subsubsection{Multi-scale Feature Fusion for Performance Improvement}

In this group of experiments, there are four cases: 1) using neither b2o nor d2e, 2) using only b2o, 3) using only d2e and 4) using both b2o and d2e. Each case is repeatedly performed three times following the experimental setting used for architecture evaluation of Stacked BNAS without KES. Moreover, the mean accuracy of three repetitive experiments is treated as the final result. Experimental results are shown in TABLE \ref{tab:fusion_ablation}.

\begin{table}[!h]
\centering
\small
\captionsetup{font={small}} 
\caption{Ablation experiments for multi-scale feature fusion of Stacked BCNN on CIFAR-10.}
\label{tab:fusion_ablation}
\begin{tabular}{ccccc}
\hline
\multirow{2}{*}{\textbf{Case}} & \multicolumn{2}{c}{\textbf{Scale Information}} & \textbf{Parameters} & \textbf{Test Error}\\
\cline{2-3}
& \textbf{d2e} & \textbf{b2o} & \textbf{(M)}  & \textbf{(\%)} \\
\hline
\hline
1 & \XSolidBrush  & \XSolidBrush  &  3.56 &  3.14 \\
2 & \Checkmark   &  \XSolidBrush  & \textbf{3.36} & 3.05 \\
3 & \XSolidBrush   & \Checkmark  &  3.64  &  2.88 \\
4 & \Checkmark  & \Checkmark & 3.70 & \textbf{2.71} \\
\hline
\end{tabular}
\end{table}

When using neither b2o nor d2e, Stacked BCNN degrades into vanilla BCNN, which lacks feature diversity for representation fusion and enhancement, so its test error is 3.14\%. For case 2, each mini BCNN employs the scale information from the broad cell to the output node and delivers 96.95\% accuracy, which is approximately 0.1\% higher than case 1. Compared with the scale information of b2o, d2e shows a greater contribution to performance improvement and obtains 97.12\% accuracy, which is approximately 0.3\% higher than case 1. In the last case, the combination of b2o and d2e further improves the performance of Stacked BCNN, i.e., 2.71\% test error. Above all, multi-scale feature fusion is effective for performance improvement of Stacked BCNN.

\section{Conclusions}
\label{sec::Conclusions}

BNASs deliver state-of-the-art efficiency via a novel broad scalable architecture named BCNN, which employs multi-scale feature fusion and hand-crafted knowledge embedding to yield satisfactory performance with shallow topology. However, there are two issues in BCNN: 1) feature diversity loss for representation fusion and enhancement and 2) time consumption of knowledge embedding design. To solve the above issues, this paper proposes Stacked BNAS. On the one hand, Stacked BNAS proposes a new broad scalable architecture named Stacked BCNN that can provide more feature diversities for representation fusion and enhancement than vanilla BCNN. On the other hand, a differentiable algorithm named KES is also proposed to learn appropriate knowledge embedding for Stacked BCNN in an automatic way instead of designing by machine learning experts. Benefiting from the combination of Stacked BCNN and an efficient optimization algorithm, the proposed Stacked BNAS delivers a state-of-the-art efficiency of 0.02 GPU days with competitive performance. Moreover, KES contributes to discovering a high-performance Stacked BCNN with fewer parameter counts. Moreover, the proposed Stacked BNAS shows powerful generalization ability on four image classification datasets. 

Nevertheless, there has also a performance gap between deep and broad search spaces in terms of accuracy on ImageNet. In the future, knowledge distillation technique will be introduced to improve the performance on ImageNet.

\bibliographystyle{IEEEtranN}
\bibliography{mybibfile}

\end{document}